\documentclass[manuscript,screen]{acmart}
\AtBeginDocument{%
  }

\setcopyright{acmlicensed}
\copyrightyear{2018}
\acmYear{2018}
\acmDOI{XXXXXXX.XXXXXXX}
\acmConference[Conference acronym 'XX]{Make sure to enter the correct
  conference title from your rights confirmation email}{June 03--05,
  2018}{Woodstock, NY}
\acmISBN{978-1-4503-XXXX-X/2018/06}

\usepackage{makecell}
\usepackage{threeparttable}
\usepackage[most]{tcolorbox}
\usepackage{multirow} 
\usepackage{comment}

\begin{document}

\title{Beyond RAG for Cyber Threat Intelligence: A Systematic Evaluation of Graph-Based and Agentic Retrieval}

\author{Dženan Hamzić}
\email{dzenan.hamzic@ait.ac.at}
\orcid{0009-0008-4698-5534}
\affiliation{%
  \institution{AIT Austrian Institute of Technology}
  \city{Vienna}
  \state{Vienna}
  \country{Austria}
}

\author{Florian Skopik}
\orcid{0000-0002-1922-7892}
\affiliation{%
  \institution{AIT Austrian Institute of Technology}
  \city{Vienna}
  \country{Austria}}
\email{florian.skopik@ait.ac.at}

\author{Max Landauer}
\orcid{0000-0003-3813-3151}
\affiliation{%
  \institution{AIT Austrian Institute of Technology}
  \city{Vienna}
  \country{Austria}
}
\email{max.landauer@ait.ac.at}

\author{Markus Wurzenberger}
\orcid{0000-0003-3259-6972}
\affiliation{%
 \institution{AIT Austrian Institute of Technology}
 \city{Vienna}
 \country{Austria}
}
\email{markus.wurzenberger@ait.ac.at}

\author{Andreas Rauber}
\orcid{0000-0002-9272-6225}
\affiliation{%
  \institution{TU Wien}
  \city{Vienna}
  \country{Austria}
}
\email{andreas.rauber@tuwien.ac.at}

\renewcommand{\shortauthors}{Hamzić et al.}

\begin{abstract}

Cyber threat intelligence (CTI) analysts must answer complex questions over large collections of narrative security reports. While retrieval-augmented generation (RAG) systems help language models access external knowledge, traditional vector-based retrieval often struggles with queries requiring reasoning over relationships between entities such as threat actors, malware, and vulnerabilities. This limitation arises because relevant evidence is often distributed across multiple text fragments and documents. Knowledge graphs address this challenge by enabling structured multi-hop reasoning over CTI data through explicit representations of entities and relationships. However, multiple retrieval paradigms, including graph-based, agentic, and hybrid approaches have emerged with differing assumptions and failure modes. It remains unclear how these approaches compare in realistic CTI settings and when graph grounding improves performance. We present a systematic evaluation of four RAG architectures for CTI analysis: standard vector retrieval, graph-based retrieval over a CTI knowledge graph, an agentic variant that repairs failed graph queries, and a hybrid approach combining graph queries with text retrieval. We evaluate these systems on 3,300 CTI question–answer pairs spanning factual lookups, multi-hop relational queries, analyst-style synthesis questions, and unanswerable cases. Our results show that graph grounding substantially improves performance on structured factual queries. The hybrid graph–text architecture improves answer quality, measured using a composite metric of agreement, adequacy, faithfulness, and clarity by up to 35\% on multi-hop questions compared to vector RAG. However, graph-only pipelines introduce failure modes such as latency variance and overconfident answers when information is missing. Architectures with query repair or hybrid retrieval achieve the most reliable overall performance.

\end{abstract}



\received{XX XX XXXX}
\received[revised]{XX XX XXXX}
\received[accepted]{XX XX XXXX}

\maketitle

\section{Introduction}

Cyber threat intelligence (CTI) plays a critical role in modern cyber defense by enabling organizations to anticipate, detect, and respond to emerging threats. By systematically collecting, analyzing, and contextualizing information about adversaries, vulnerabilities, and attack campaigns, CTI supports informed decision-making and proactive security strategies. As cyber threats continue to grow in scale and sophistication, effective CTI analysis has become essential for maintaining situational awareness and reducing organizational risk. CTI analysts must continuously distill long-form narrative reports into actionable insights about threat actors, malware families, infrastructure vulnerabilities, and attacker campaigns.
Large language models (LLMs) offer a natural-language interface that allows analysts to query CTI reports directly, but in operational CTI the dominant risk is decision error: confident but incorrect answers that can lead to incorrect response prioritization, delay mitigation, or distort situational awareness.
Prompting without external knowledge grounding is therefore insufficient in high-stakes settings because LLMs may hallucinate entities, relations, and supporting evidence with high confidence \cite{ji2022hallucination}.

\begin{figure}[ht]
    \centering
    \includegraphics[width=0.6\linewidth]{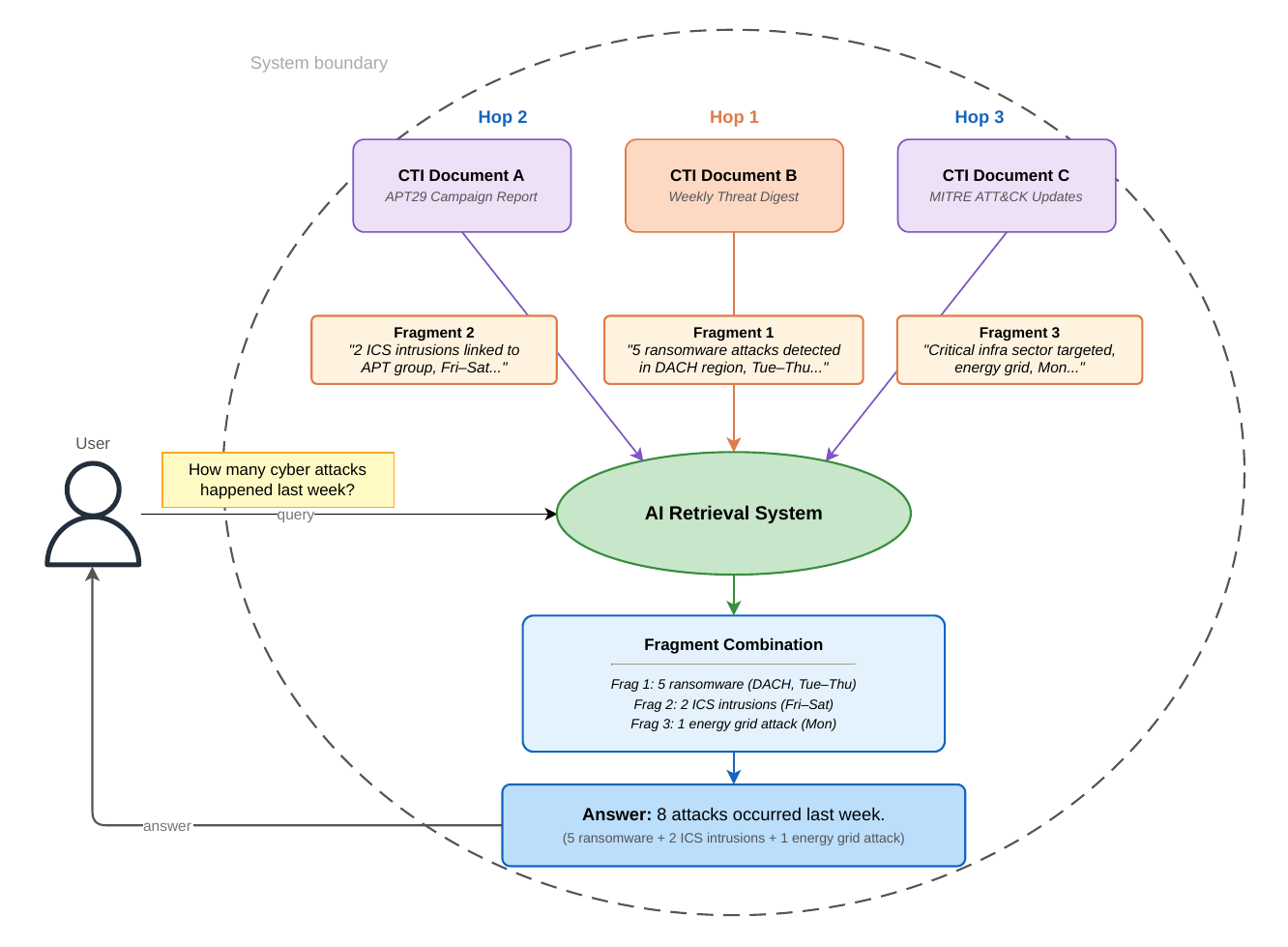}
    \caption{AI Retrieval in CTI Domain.}
    \label{fig:intro_rag_architecture}
    \vspace{-3mm}
\end{figure}

Retrieval-Augmented Generation (RAG) mitigates this risk by grounding responses in external evidence retrieved at inference time \cite{lewis2020rag,gao2024rag_survey}.
Most deployed systems implement dense retrieval over chunked text~\cite{karpukhin2020dpr}, where queries and documents are encoded into continuous vector representations and matched via similarity search (e.g., using approximate nearest-neighbor methods such as FAISS~\cite{johnson2017faiss}), followed by generation conditioned on the retrieved snippets.
However, CTI questions are frequently relational and temporal, requiring multi-hop reasoning, where answering a query involves chaining together multiple related facts across entities, CTI reports, or time (e.g., actor $\rightarrow$ uses $\rightarrow$ malware $\rightarrow$ targets $\rightarrow$ sector, or comparing campaigns over time).
Dense retrieval that returns the top-$k$ most relevant text chunks~\cite{karpukhin2020dpr, lewis2020rag} can fail when evidence is distributed across distant text fragments, when constraints must be satisfied jointly, or when the answer depends on chaining multiple facts \cite{yang2018hotpotqa}.
Equally important, LLM-based CTI assistants must reliably abstain when the reports do not support an answer, rather than filling gaps with plausible speculation \cite{rajpurkar2018know}. Accordingly, we treat unsafe abstention, i.e., the failure to correctly
identify unanswerable queries, and latency instability, i.e., unpredictable
response times in time-critical settings, as security-relevant failure modes in CTI assistants.

Graph-based retrieval has re-emerged as a promising direction: converting unstructured CTI text into a property graph representation~\cite{angles2018propertygraph} enables explicit grounding of entities and relationships and allows queries to traverse these relations under structural constraints, facilitating multi-hop reasoning over interconnected threat intelligence. Yet, structure introduces new operational failure modes. Answering now depends on generating and executing correct graph queries (e.g., Cypher, a declarative query language for graphs \cite{francis2018cypher}), and failures may arise from text-to-query translation errors, schema mismatches, or empty-result queries. All these issues could erase the theoretical benefits of graph grounding~\cite{xiang2025whentouse, liang2025graphragunderfire}.
Moreover, partial structure can produce overconfident answers when evidence is incomplete, and iterative correction loops can introduce large and unpredictable latency, both security-relevant risks in time-critical CTI workflows.
Recent agentic patterns suggest a remedy: plan-act-reflect loops can detect and repair tool-use mistakes during execution \cite{yao2023react,shinn2023reflexion}.
Still, the community lacks controlled evidence for when graph grounding helps on realistic CTI workloads, how much correction is needed for reliability, and what runtime-quality trade-offs arise when moving beyond RAG.


This paper provides a controlled evaluation of four representative retrieval 
architectures for CTI question answering, selected to span core design 
dimensions in retrieval-augmented generation.

(i) \textbf{Semantic RAG (RAG)} over text chunks follows the standard 
dense-retrieval paradigm introduced by Lewis et al.~\cite{lewis2020rag}, 
relying on local semantic similarity search under context-window constraints.

(ii) \textbf{Graph-only retrieval (GRAG)} replaces local chunk retrieval with 
structured text-to-Cypher translation over a property graph. This design supports 
global structural reasoning and aggregation in the spirit of recent 
knowledge-graph–augmented RAG systems~\cite{pan2024unifying,edge2025graphrag,zhu2025kg2rag}.

(iii) \textbf{Agentic GRAG (AGRAG)} extends GRAG with critique-and-repair 
loops inspired by agentic tool-use and self-reflection frameworks such as 
ReAct and Reflexion~\cite{yao2023react,shinn2023reflexion}, mitigating 
query-generation brittleness in structured pipelines.

(iv) \textbf{HybridRAG (HRAG)} combines structured graph querying with 
unstructured semantic retrieval, reflecting hybrid knowledge–vector 
integration strategies proposed in recent multimodal and KG-RAG systems 
\cite{papageorgiou2025multimodal,pan2024unifying,10.1007/978-3-032-00633-2_3}.

These architectures do not cover all RAG variants, but represent 
distinct retrieval paradigms discussed in the literature: dense 
retriever-based RAG \cite{lewis2020rag,karpukhin2020dpr}, structured 
retrieval over knowledge graphs \cite{sun-etal-2019-pullnet}, agentic retrieval 
with iterative query repair \cite{yao2023react,shinn2023reflexion}, and 
hybrid systems combining structured and semantic retrieval. Other variants such as reranking, adaptive retrieval routing,
or multimodal RAG have also been explored \cite{nogueira2019passage,asai2019learning,gao2023retrieval}.
These extensions mainly modify components within these architectural families.

We evaluate the four RAG systems using five different LLMs on 3{,}300 automatically generated question-answer pairs spanning \emph{simple}, \emph{single-hop}, \emph{multi-hop}, \emph{guided analyst-style}, and \emph{unanswerable} questions.
Beyond answer quality, we explicitly analyze security-relevant failure modes, unsafe abstention behavior and latency instability, showing that naive graph-only retrieval can underperform RAG unless paired with correction and redundancy mechanisms.

Our contributions are summarized as follows:
\begin{itemize}
\item \textbf{A CTI evaluation dataset and generation pipeline.} We contribute a dataset of 3{,}300 labeled QA pairs with controlled question types and query provenance, enabling reproducible evaluation of CTI-focused RAG variants.
\item \textbf{A systematic, controlled comparison.} We compare semantic RAG, graph-only RAG, agentic GraphRAG, and hybrid graph-text retrieval under a shared CTI knowledge base and a unified evaluation protocol.
\item \textbf{A systematic analysis of graph-based retrieval trade-offs.} We quantify per-question-type benefits and show that naive graph-only retrieval can degrade safety on unanswerable queries and induce runtime instability, motivating explicit safeguards.
\item \textbf{Security-relevant failure and robustness analysis.} We analyze how structured retrieval changes CTI failure modes, including overconfident answers from partial structure, cascading errors from structured backends, and latency spikes from iterative query repair-properties that directly affect the trustworthiness of CTI assistants.
\end{itemize}

\label{subsec:research_questions}

To systematically analyze the effectiveness, robustness, and cost of graph-based retrieval for cyber threat intelligence, we investigate the following research questions:

\begin{itemize}
  \item \textbf{RQ1:} To what extent do graph-based and hybrid RAG systems improve answer quality over semantic RAG on CTI tasks?
  \item \textbf{RQ2:} To what extent do different question types (simple, single-hop, multi-hop, guided, unanswerable) benefit from explicit graph grounding?
  \item \textbf{RQ3:} How does the underlying LLM affect text-to-Cypher generation performance?
  \item \textbf{RQ4:} How large is the runtime-quality trade-off across systems and models?
\end{itemize}

The remainder of this paper is structured as follows. Section~\ref{sec:background} reviews related work on retrieval-augmented generation, graph-based retrieval, and agentic and hybrid RAG systems, positioning our contribution within prior research. Section~\ref{sec:methodology} describes the CTI dataset, graph construction process, experimental design, and evaluation methodology. Section~\ref{sec:results} presents quantitative results across retrieval architectures, question types, and language models, including robustness and runtime analyses. Section~\ref{sec:discussion} discusses the implications of these findings, with a focus on security-relevant failure modes and deployment trade-offs. Finally, Section~\ref{sec:conclusion} concludes the paper and outlines directions for future work.

\section{Background and Related Work}
\label{sec:background}
This section reviews the foundations of retrieval-augmented generation and related work on graph-based and agentic retrieval architectures relevant to our study.

\subsection{Retrieval-Augmented Generation}
RAG combines parametric language models with external knowledge sources retrieved at inference time, enabling
models to ground responses in supporting evidence and reduce hallucinations~\cite{lewis2020rag}.
In a standard RAG pipeline, a user query is embedded, a retriever selects the top-k relevant text passages using dense
retrieval technique (e.g. Dense Passage Retrieval)~\cite{karpukhin2020dpr}, and the language model generates an answer
conditioned on the retrieved context. However, several limitations have been identified. In particular, top-k passage retrieval often fails when relevant evidence is
distributed across multiple documents or when answering requires multi-hop reasoning across entities and relations~\cite{yang2018hotpotqa}.

\subsection{Graph–Augmented Retrieval Generation}
One direction for improving RAG systems is the integration of knowledge graphs or property graphs that explicitly represent
entities and relationships. In such systems, unstructured text is converted into graph representations (using e.g. LLM prompting),
enabling queries to traverse structured relationships and aggregate evidence across entities~\cite{edge2025graphrag,pan2024unifying,zhu2025kg2rag}.

Graph-based retrieval can improve performance on relational queries and multi-hop reasoning tasks by explicitly following
entity relationships. However, benefits of graph grounding may depend strongly on query complexity and graph coverage \cite{xiang2025whentouse}.

\subsection{Agentic and Self-Correcting Retrieval}

Recent work explores agentic RAG architectures that augment retrieval pipelines with iterative reasoning and tool use.
Frameworks such as ReAct~\cite{yao2023react} and Reflexion~\cite{shinn2023reflexion} demonstrate that language models can
improve task performance by alternating between reasoning steps, tool execution, and self-reflection. 
In retrieval pipelines, these mechanisms enable systems to detect and repair failures during query generation or tool interaction.

\subsection{GraphRAG Applications and Evaluation Challenges}

\begin{table*}[t]
\centering
\small
\caption{Summary of related graph-based RAG work}
\label{tab:related_work_summary}
\begin{tabular}{p{2.7cm}p{2.5cm}p{8cm}}
\toprule
\textbf{Paper} & \textbf{Architecture focus} & \textbf{Relevance to this work} \\
\midrule
Edge et al.\ \cite{edge2025graphrag} & Hierarchical GraphRAG & Hierarchical graph retrieval for summarization; motivates multi-level retrieval \\
Pan et al.\ \cite{pan2024unifying} & KG--LLM integration & Conceptual taxonomy for graph–LLM integration \\
Wu et al.\ \cite{wu2025medicalgraphrag} & Domain GraphRAG & Medical GraphRAG with domain-specific enhancements \\
Agentic Medical \cite{agentic_clinical_rag2025} & Agentic GraphRAG & Self-correcting agentic refinement for structured retrieval \\
Papageorgiou et al.\ \cite{papageorgiou2025multimodal} & Hybrid GraphRAG & Hybrid multi-agent GraphRAG; explainability benefits \\
Liang et al.\ \cite{liang2025graphragunderfire} & Robustness analysis & Security evaluation of GraphRAG under poisoning attacks \\
\bottomrule
\end{tabular}
\vspace{-5mm}
\end{table*}

Table~\ref{tab:related_work_summary} summarizes the related work. Graph-based RAG (GraphRAG) incorporates knowledge graphs to enable entity-centric reasoning and relational traversal \cite{pan2024unifying, zhang2025whentouse, zhu2025kg2rag}. Edge et al.\ demonstrate hierarchical GraphRAG for query-focused summarization \cite{edge2025graphrag}, while Xiang et al.\ show GraphRAG benefits depend critically on query complexity \cite{xiang2025whentouse}. 

Domain-specific applications illustrate both opportunities and challenges: medical GraphRAG systems \cite{wu2025medicalgraphrag,agentic_clinical_rag2025} show improved safety through structured search, while hybrid approaches \cite{papageorgiou2025multimodal} leverage multiple retrieval modalities to compensate for individual weaknesses. However, Liang et al.\ demonstrate that GraphRAG's relational structure creates novel attack surfaces, with poisoning propagating across interconnected entities \cite{liang2025graphragunderfire}, underscoring the need to jointly evaluate quality and robustness. Prior work reports heterogeneous and task-dependent outcomes for graph-/KG-augmented RAG, and evaluation practices remain inconsistent across pipelines, tasks, and domains, which complicates direct cross-paper comparison~\cite{peng2025graphrag}. Consequently, many studies introduce and test a single proposed variant on one or a small set of benchmarks, limiting isolation of causal factors~\cite{zhu2025kg2rag}.

Our work addresses this gap through a controlled evaluation of four retrieval architectures in the CTI domain. By holding data, models, and evaluation constant, we disentangle the effects of explicit graph grounding, agentic query refinement, and hybrid retrieval redundancy.

\section{Methodology}
\label{sec:methodology}

\begin{figure*}
    \centering
    \includegraphics[width=1\linewidth]{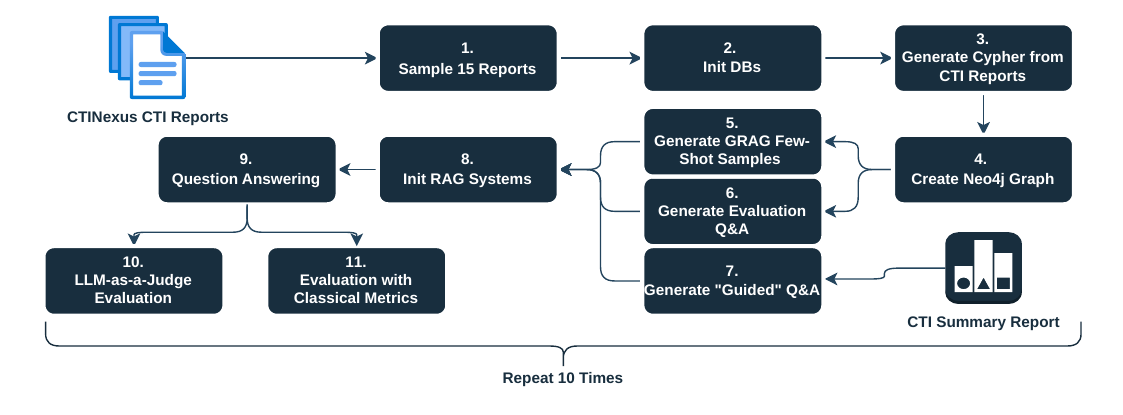}
    \caption{Data generation and evaluation workflow.}
    \label{fig:eval-workflow}
    \vspace{-5mm}
\end{figure*}

The goal of this study is to systematically evaluate how different retrieval architectures affect the reliability and quality of LLM-based question answering over CTI reports. We compare four retrieval paradigms: RAG, GRAG, AGRAG, and HRAG. These architectures represent different approaches to grounding LLM responses in external knowledge sources and exhibit distinct trade-offs between contextual coverage, relational reasoning capabilities, and robustness to missing information.

To enable a controlled comparison, we construct an evaluation framework that transforms unstructured CTI reports into structured textual representations. First, CTI documents are converted into Cypher statements using an LLM-based text-to-graph extraction process. These statements populate a Neo4j property graph that captures entities such as threat actors, malware families, vulnerabilities, and their relationships. 

Question–answer pairs are derived from the generated Cypher queries. Generating questions from Cypher ensures that the evaluation only tests information that is verifiably present in the knowledge graph and corresponding CTI reports. The resulting dataset contains multiple question types, including simple lookups, relational queries, multi-hop reasoning tasks, analyst-style guided questions, and unanswerable cases.

Each RAG architecture then answers the generated questions using identical inputs and underlying knowledge sources. System outputs are evaluated using both classical automatic metrics and an LLM-as-a-Judge framework to assess answer quality, hallucination behavior, abstention capability, and runtime stability. The following subsections describe the formal problem formulation, the evaluation framework, dataset construction, retrieval architectures, and evaluation metrics in detail.

\subsection{Formal Problem Definition}

We model CTI question answering as a retrieval-augmented generation task over a document corpus and an associated knowledge graph. Let 
$\mathcal{D} = \{d_1, d_2, \dots, d_n\}$ denote a corpus of CTI reports. From this corpus, a knowledge graph is constructed and represented as 
$G = (V, E)$, where $V$ denotes entities (e.g., Malware, ThreatActor, Tool, Victim) and $E$ represents relations (e.g., uses, targets, exploits) between these entities.

A set of evaluation questions is defined as 
$Q = \{q_1, q_2, \dots, q_m\}$, where each question $q_i$ is associated with a reference answer $a_i$. For a given question $q_i$, a retrieval architecture $R_k$ (e.g., graph retrieval, dense retrieval, or both) retrieves a set of contextual evidence $C_k$:

\[
R_k(q_i) \rightarrow C_k
\]

The retrieved context $C_k$ is then provided to a LLM model (A) that generates an answer $\hat{a}_{k,i}$:

\[
A(q_i, C_k) \rightarrow \hat{a}_{k,i}
\]

We evaluate four retrieval architectures $k \in \{\text{RAG}, \text{GRAG}, \text{AGRAG}, \text{HRAG}\}$, which differ in how contextual evidence $C_k$ is constructed.
\begin{itemize}

\item \textbf{RAG.}
The retrieval function $R_{\text{RAG}}$ retrieves a set of semantically relevant document passages from $\mathcal{D}$ using vector similarity search or dense search.

\item \textbf{GRAG.}
The retrieval function $R_{\text{GRAG}}$ translates the question $q_i$ into a graph query (Cypher) over $G$, retrieving relational evidence in the form of entity subgraphs.

\item \textbf{AGRAG.}
The retrieval function $R_{\text{AGRAG}}$ extends graph retrieval with an iterative critique-and-repair mechanism that refines graph queries.

\item \textbf{HRAG.}
The retrieval function $R_{\text{HRAG}}$ combines semantic document retrieval and graph traversal, producing a hybrid context set containing both textual passages and graph-derived relational evidence.

\end{itemize}
The evaluation compares architectures with respect to answer quality, hallucination behavior, abstention performance, and runtime stability by measuring agreement between generated answers $\hat{a}_{k,i}$ and reference answers $a_i$ across the question set $Q$.

The following subsections describe how the Q\&A dataset, knowledge graph, RAG parameters, retrieval pipelines, and evaluation metrics are constructed and configured within this framework.

\subsection{The Evaluation Framework}

Because LLM outputs are probabilistic, we perform multiple evaluation runs instead of relying on a single experiment. We conduct five independent evaluations using different LLMs, each consisting of ten runs. In each run, 15 CTI reports are randomly sampled from the CTINexus dataset~\cite{Cheng2025_official}, translated into Cypher queries, and inserted into the graph database. Figure~\ref{fig:eval-workflow} illustrates the overall experimental workflow.

After random file selection in Step~1, in order to have a fair and unbiased evaluation across the runs, both the FAISS\footnote{\url{https://github.com/facebookresearch/faiss}} vector database and the Neo4j\footnote{\url{https://neo4j.com/product/neo4j-graph-database/}} graph database are reset in Step~2. In Step~3, the unstructured CTI text is converted into Neo4j Cypher statements using a LLM (see Section~\ref{sect:Graph_Structure}). In Step~4, the generated Cypher statements are executed to populate the graph database. If an insertion error occurs, the Cypher query and the corresponding error message are passed back to the LLM for correction, and the insertion is retried.

Once all CTI texts have been successfully converted into Cypher, Step~5 generates few-shot question-to-Cypher examples, which are used to parametrize the GRAG architecture, using an LLM-based prompt\footnote{\url{https://github.com/ait-cti/beyond-vanilla-rag/blob/main/grag-few-shot-qc-samples.md}}. These generated examples, (question, query) pairs, are subsequently used as few-shot examples during GraphRAG inference. The prompt instructs the model to produce a fixed number of (question, query) pairs in strict JSON format, ensuring deterministic parsing and reproducibility. It enforces multiple structural constraints:

\begin{itemize}
    \item \textbf{Read-only Cypher restriction.} Only non-mutating clauses are permitted. All write operations are explicitly forbidden to prevent unsafe query generation.
    
    \item \textbf{Schema grounding.} The model is required to use only labels, relationship types, and properties defined in the authoritative graph schema\footnote{\url{https://github.com/ait-cti/beyond-vanilla-rag/blob/main/text-to-cypher-prompt.md}}, used in Step~3, and the previously generated Cypher inserts. This prevents hallucinated entity types or relations.
    
    \item \textbf{Executability guarantees.} Queries must be directly runnable without any additional parameters, avoid hard-coded values not present in the graph, and return minimal, clearly aliased outputs.
    
    \item \textbf{Diversity constraints.} The prompt enforces a balanced mix of question types, including simple node lookups, one-hop traversals, multi-hop paths, and aggregate queries, ensuring coverage of different reasoning patterns.
    
    \item \textbf{Self-validation checks.} Before returning the output, the model is instructed to silently verify JSON validity, item count correctness, and clause compliance.
\end{itemize}

These constraints ensure that the generated few-shot examples are syntactically valid, schema-consistent, and representative of the reasoning patterns required for GRAG evaluation. For each incoming user question,
the system retrieves the top-k most semantically similar example pairs and injects them into the prompt before Cypher generation. This in-context learning setup guides the LLM to produce schema-compliant, executable Cypher queries by imitating the structure and patterns of the provided examples. 

In Step~6, evaluation questions are generated from the Cypher queries using LLM prompting\footnote{\url{https://github.com/ait-cti/beyond-vanilla-rag/blob/main/cypher-to-qa-pairs.md}}. Deriving questions from executable Cypher queries serves as a control to ensure that answerability is determined by the shared knowledge base across all RAG architectures. Importantly, this does not provide graph-based systems with privileged information: all systems receive only the natural-language question at inference time, and RAG has access to the same underlying facts through the original CTI texts. Performance differences therefore reflect retrieval and reasoning effectiveness. In this step, four types of questions are generated from the CTI Cypher queries: 15 simple, 15 single-hop, 15 multi-hop, and 5 unanswerable questions. These categories follow established distinctions in question answering (QA) research and are particularly suited for evaluating structured retrieval. Simple and single-hop questions correspond to entity lookup and one-relation reasoning in knowledge-graph QA and semantic parsing, requiring retrieval of node properties or traversal of a single edge. Multi-hop questions require compositional reasoning across multiple relations, consistent with multi-hop benchmarks such as HotpotQA~\cite{yang2018hotpotqa}, and test the system’s ability to combine distributed evidence. Unanswerable questions follow the SQuAD~2.0 paradigm \cite{rajpurkar2018know}, requiring models to detect insufficient evidence and abstain rather than hallucinate, an essential property in CTI settings. We exclude formats such as multiple-choice or opinion-based questions, as they do not reflect realistic CTI workflows.

In Step~7, we introduce an additional set of 16 ``guided'' questions derived from the Australian Government's Annual CTI Report\footnote{\url{https://github.com/ait-cti/beyond-vanilla-rag/blob/main/Annual_Cyber_Threat_Report_2024-25.pdf}} 
to approximate realistic cybersecurity analysis tasks. Unlike the Cypher-derived categories (simple, single-hop, multi-hop, unanswerable), guided questions originate from an external CTI document and are therefore not guaranteed by graph construction or schema coverage. An LLM is prompted\footnote{\url{https://github.com/ait-cti/beyond-vanilla-rag/blob/main/guided_qa_generation.md}} to generate 16 questions (8 multi-hop and 8 of other types, like simple or single-hop questions) covering 
strategic CTI themes such as threat actors, TTPs, sectors, and geopolitical context. To reduce bias and triviality, generated questions are validated for answerability against the selected CTINexus reports, filtered to remove 
near-trivial fact lookups, and required to reference entities present in the corpus. This design reflects analyst-driven intelligence requirements, where questions emerge from external reporting needs rather than schema-constrained database exploration.

In Step~8, the vector-based RAG database is initialized. The CTI texts are chunked and embedded into the vector database, and the HybridRAG system is initialized by embedding textual representations of entities and relationships using OpenAI’s text-embedding-3-large\footnote{\url{https://platform.openai.com/docs/models/text-embedding-3-large}} model.

In Step~9, the evaluation phase begins, during which all systems answer a total of 66 questions per run. Finally, in Steps~10 and~11, system outputs are evaluated using classical metrics such as F1, BLEU, and ROUGE, complemented by an LLM-as-a-Judge evaluation\footnote{\url{https://github.com/ait-cti/beyond-vanilla-rag/blob/main/llm-judge.md}}~\cite{Gu2025} applied to the generated answers.

\subsection{Generated Q\&A Validation}
Table~\ref{tab:question_quality} validates question generation quality across three dimensions. First, pattern diversity: 76\% of questions begin with unique four-word prefixes, indicating the generation process avoids repetitive phrasing. Second, answer conciseness: for factoid categories (simple, single-hop, multi-hop, unanswerable), which require precise retrievable answers, 99.4\% of gold answers fall within 12 words, what aligns with one of the requirements in the Q\&A generation prompt, consistent with answer-length distributions observed in established QA benchmarks such as SQuAD~\cite{rajpurkar2016squad}(most answers are short spans, with typical lengths below 10 tokens) and Natural Questions~\cite{kwiatkowski2019natural}(answers similarly concentrated in short spans, generally under $\sim$10 tokens). Guided questions are excluded from this constraint by design, as they elicit multi-sentence explanations that test reasoning chains rather than factoid retrieval. Third, compositional coverage: each run contains at least 5 aggregate multi-hop questions. Across runs, the number of such questions per run has a mean of 7.2 (range: 5–9), providing sufficient per-run sample size to compute meaningful sub-category scores for compositional reasoning. Questions reference entities present in the constructed knowledge graph, ensuring the evaluation tests retrieval of real CTI concepts. This design guarantees a fair comparison between graph-based and semantic retrieval, as the evaluated knowledge is shared between both representations.

\begin{table}[t]
\centering
\caption{Generated Question Quality and Validation Metrics}
\label{tab:question_quality}
\begin{tabular}{lr}
\toprule
Metric & Value \\
\midrule
\multicolumn{2}{l}{\textbf{Question Diversity}} \\
Total questions (per run) & 66 \\
Unique 4-word prefixes & 398/522 (76.1\%) \\
\midrule
\multicolumn{2}{l}{\textbf{Answer Conciseness (factoid categories only)}} \\
Categories included & simple, single/multi-hop, unans. \\
Answers $\leq$12 words & 99.4\% \\
Mean answer length (words) & 1.6 \\
Median answer length (words) & 1 \\
\midrule
\multicolumn{2}{l}{\textbf{Guided Answers (excluded from constraint)}} \\
Mean answer length (words) & 39.4 \\
\midrule
\multicolumn{2}{l}{\textbf{Aggregate Multi-hop Coverage}} \\
Aggregates per run (mean $\pm$ sd) & 7.2 $\pm$ 1.4 \\
Range across runs & 5--9 \\
Runs with $\geq$5 aggregates & 10/10 \\
\bottomrule
\end{tabular}
\vspace{-5mm}
\end{table}

\subsection{RAG Systems Configuration}
This section describes the configuration of the 4 RAG systems.
\begin{figure*}
    \centering
    \includegraphics[width=1.0\linewidth]{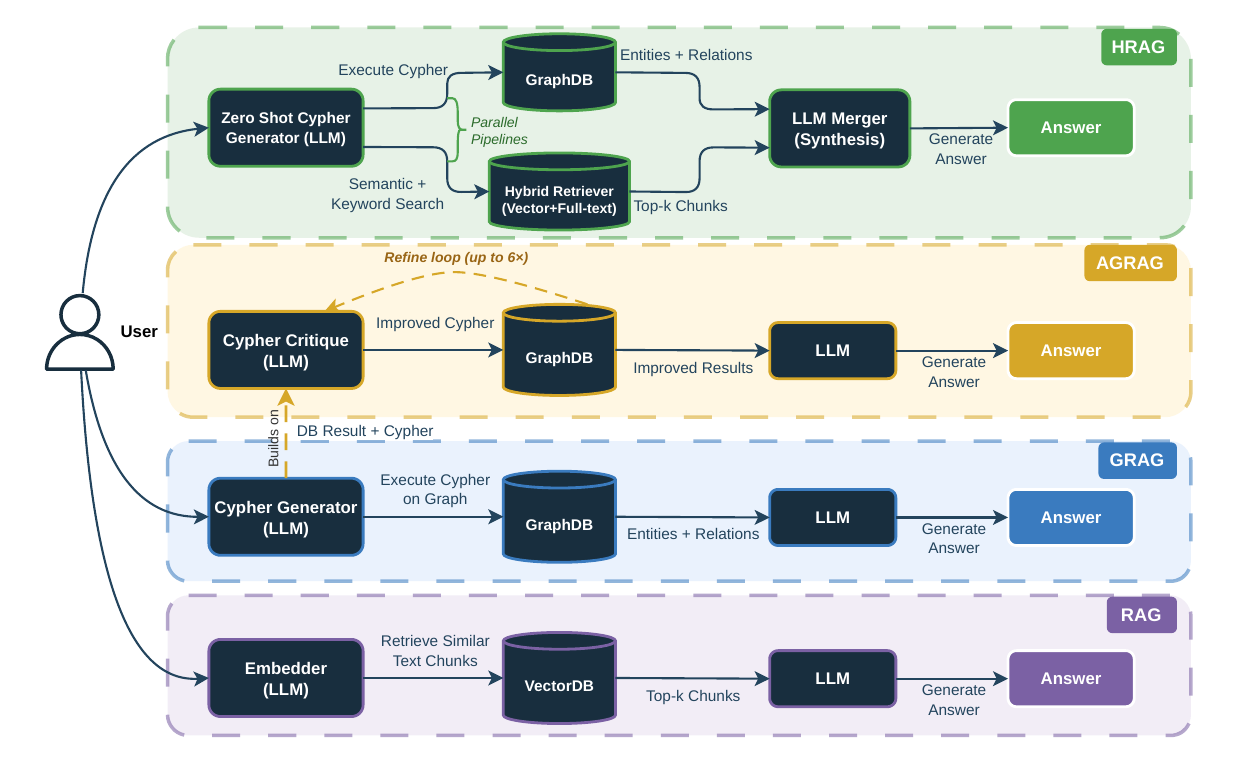}
    \caption{Overview of different RAG systems.}
    \label{fig:rag_systems}
    \vspace{-5mm}
\end{figure*}

\subsubsection{RAG}
LLMs have a limited prompt context window, which prevents the entire contents of a database from being injected during the retrieval phase. To reflect this constraint, we limit the number of retrieved text chunks to $k=3$, meaning that only the three most semantically similar chunks are inserted into the RAG prompt. The chunk size is set to 200 characters with an overlap of 20 characters~\cite{wang-etal-2025-document,jiang2024longragenhancingretrievalaugmentedgeneration}.

This configuration prevents overloading the model’s context window, which could otherwise lead to degraded performance or catastrophic forgetting~\cite{huang-etal-2025-selfaug}. As illustrated in Figure~\ref{fig:rag_systems}, the user query is first embedded, after which the top-$k$ similar chunks are retrieved from the vector database and provided to the LLM. The LLM then generates the final answer conditioned on both the user query and the retrieved context, following a predefined RAG prompt\footnote{\url{https://github.com/ait-cti/beyond-vanilla-rag/blob/main/RAG-prompt.md}}.

\subsubsection{GRAG}

\begin{figure*}
    \centering
    \includegraphics[width=1\linewidth]{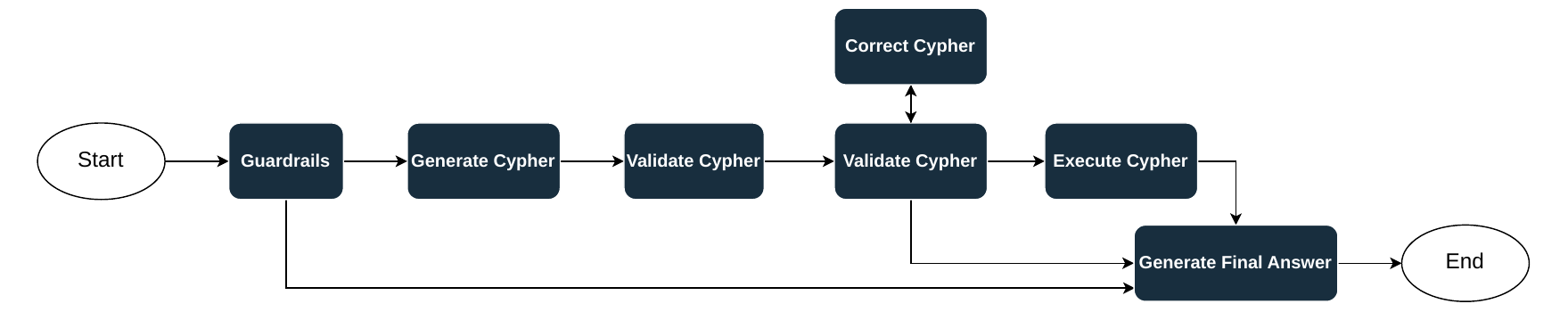}
    \caption{GRAG system configuration.}
    \label{fig:GRAG}
    \vspace{-4mm}
\end{figure*}

As illustrated in Figure~\ref{fig:rag_systems}, GRAG uses a graph database (Neo4j) as the primary source for information retrieval. As shown in Figure~\ref{fig:eval-workflow}, in Step~5, few-shot question-to-Cypher example pairs are generated and stored. These pairs serve as reference examples for translating natural-language queries into Cypher statements.

When a user submits a query, the system retrieves the three most similar (question, Cypher) pairs and inserts them as few-shot examples into the prompt. Cypher generation then follows an iterative control loop (Figure~\ref{fig:GRAG}).

First, guardrails perform a lightweight domain check and reject non-CTI queries before graph interaction. If accepted, the LLM is prompted\footnote{\url{https://github.com/ait-cti/beyond-vanilla-rag/blob/main/GRAG-prompts.md}} with the user question, Neo4j schema, and retrieved few-shot examples to generate a candidate read-only Cypher query. The query is validated for syntactic correctness and schema conformity and executed against Neo4j to detect runtime or empty-result errors. If validation fails, the error message and previous query are fed back to the LLM for repair. This generate–validate–repair loop continues until successful execution or a maximum of 25 iterations is reached. Once execution succeeds, the database results are converted into a natural-language answer. The implementation is based on LangChain\footnote{\url{https://www.langchain.com/}} and LangGraph\footnote{\url{https://www.langchain.com/langgraph}}, following the reference architecture provided by the LangChain framework.

\subsubsection{AGRAG}
AGRAG builds on the output of GRAG, namely the natural-language query, the generated Cypher code, and the database results. It employs an LLM to \emph{comment on and critically assess} the generated Cypher query (see Figure~\ref{fig:rag_systems}) using a dedicated critique prompt\footnote{\url{https://github.com/ait-cti/beyond-vanilla-rag/blob/main/AGRAG-prompt.md}}.

Given the user question and the Neo4j database schema, AGRAG first evaluates the Cypher query produced by GRAG. If issues are identified, the LLM returns a refined version of the Cypher query together with an explanatory comment. The refined query is then executed against the database. In the presence of execution errors or remaining inconsistencies, the critique-and-refinement loop is repeated for up to six iterations. In practice, depending on the underlying LLM, this iterative refinement is rarely required more than once, as most Cypher errors are resolved during the first correction step.

\subsubsection{HRAG}

\begin{figure}
    \centering
    \includegraphics[width=0.6\linewidth]{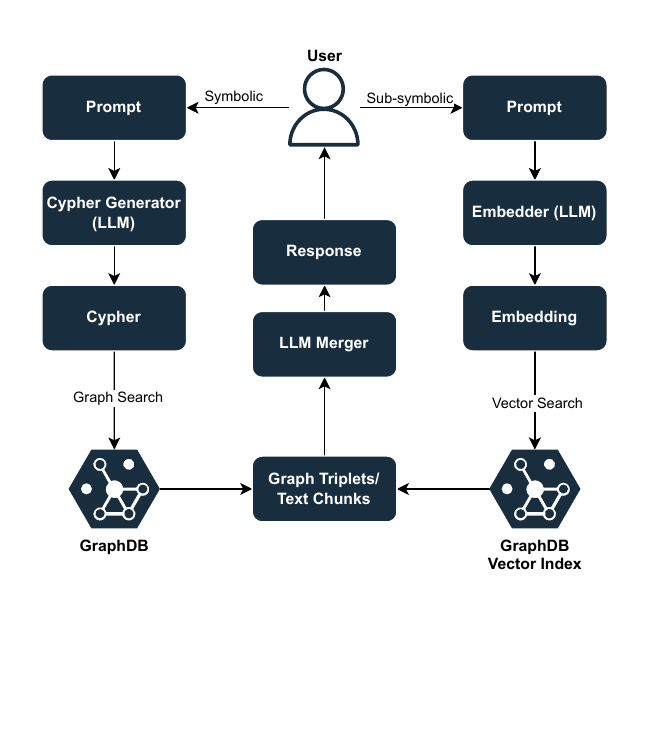}
    \vspace{-25.0mm}
    \caption{HybridRAG parallel pipeline}
    \label{fig:hrag_pipeline}
    \vspace{-6mm}
\end{figure}

HybridRAG builds on the insight from prior work~\cite{papageorgiou2025multimodal,pan2024unifying,10.1007/978-3-032-00633-2_3} that combining symbolic knowledge graphs with semantic vector search enables complex reasoning over CTI data that neither modality achieves alone. This system first creates a unified search layer inside Neo4j called \texttt{SearchDoc}. 
All relevant graph nodes (e.g., ThreatActor, Malware, Victim) and relationships (e.g., uses, attacked, exploits) 
(see Sect.~\ref{sect:Graph_Structure} for the complete schema) are converted into short, human-readable textual representations and stored in a unified text property that aggregates the descriptive content of each Neo4j element. This allows the system to search the entire graph in a consistent way, regardless of whether the information originally came from a node or an edge. On top of this SearchDoc layer, two indexes are created:
\begin{itemize}
  \item a full-text index for keyword-based search (names, IDs, exact terms),
  \item a vector index for semantic search using embeddings (search by meaning).
\end{itemize}
A \textbf{hybrid retriever} combines both approaches, so each question retrieves relevant SearchDoc entries using keyword matching and semantic similarity together. At query time, HybridRAG runs two pipelines (see Fig.~\ref{fig:hrag_pipeline}) in parallel:
\begin{enumerate}
  \item Graph pipeline (Symbolic): an LLM translates the question into a Cypher query (Zero Shot Generator in Figure~\ref{fig:rag_systems}), executes it on Neo4j, 
        and returns structured results such as tables, counts, and explicit relationships. 
        If the query fails, simple rule-based fixes are applied first, followed by LLM-based repair if needed.
  \item Retrieval pipeline (Sub-Symbolic): the hybrid retriever fetches unstructured SearchDoc text snippets 
        that provide descriptive and contextual information.
\end{enumerate}

Finally, a synthesis LLM prompt\footnote{\url{https://github.com/ait-cti/beyond-vanilla-rag/blob/main/HRAG-prompt.md}} merges both outputs. It prioritizes the graph results for exact facts and relationships, while using the retrieved text to enrich the answer or fill gaps. The final response is concise, consistent, and grounded in the underlying graph data.

\subsection{Dataset}\label{dataset_desc}
In this paper, we use the publicly available CTINexus~\cite{Cheng2025_official} dataset, which consists of 150 CTI reports collected from well-recognized threat-sharing platforms. We select the CTINexus dataset because it offers a large-scale collection of real-world CTI reports written in complex, domain-specific language, enabling a realistic evaluation of retrieval-augmented generation methods on unstructured security texts. The authors of CTINexus performed entity and relationship extraction on the original reports, resulting in 26 unique cybersecurity-related entity types and 855 unique cybersecurity-related relationships connecting these entities across the 150 reports.

\subsection{Generated Graph Analysis and Validation}
We confirm the structural validity of the resulting knowledge graph (from unstructured CTI texts via LLM prompt to Cypher) by analyzing the structural properties of the generated graph across runs. Table~\ref{tab:graph_stats} summarizes the graph structure generated from the CTINexus reports for 10 runs within a single evaluation. Text-to-graph conversion produces an average of 10.9 entities and 1.6 relationships per report through 104 lines of Cypher code with 23.4 MERGE operations, where MERGE is a Cypher clause that creates or matches nodes and relationships to ensure non-duplicative graph construction, demonstrating substantial structural complexity beyond simple entity catalogs.


\begin{table*}[t]
\centering

\begin{minipage}[t]{0.38\linewidth}
\centering
\caption{CTI Knowledge Graph Statistics}
\label{tab:graph_stats}
\begin{tabular}{lr}
\toprule
Metric & Value \\
\midrule
Total reports (across 10 runs) & 150 \\
Unique report files & 70 \\
Avg entities per report & 10.9 $\pm$ 5.5 \\
Avg relationships per report & 1.6 $\pm$ 2.4 \\
Avg Cypher lines per report & 104 $\pm$ 54 \\
Avg MERGE statements & 23.4 $\pm$ 12.6 \\
Unique entity types & 17 \\
Unique relationship types & 8 \\
Avg report word count & 137 \\
\bottomrule
\end{tabular}
\end{minipage}
\hfill
\begin{minipage}[t]{0.60\linewidth}
\centering
\caption{Schema Validation and Evidence Attribution Metrics (aggregated over 10 runs).}
\label{tab:schema_validation}
\begin{tabular}{lrr}
\toprule
Metric & Value & Compliance \\
\midrule
\multicolumn{3}{l}{\textbf{Schema Compliance}} \\
Entity type compliance         & 1{,}044/1{,}044 & 100.0\% \\
Relationship type compliance  & 438/438         & 100.0\% \\
Country code format (ISO 3166-1 $\alpha$-2) & 30/30      & 100.0\% \\
\midrule
\multicolumn{3}{l}{\textbf{Evidence Attribution}} \\
Relationships with \texttt{evidence}     & 178/438 & 40.6\% \\
Relationships with \texttt{source\_id}   & 102/438 & 23.3\% \\
Relationships with \texttt{page}         &  89/438 & 20.3\% \\
\midrule
\multicolumn{3}{l}{\textbf{CVE Extraction Recall}} \\
CVEs in source text extracted as nodes   & 63/63   & 100.0\% \\
\bottomrule
\end{tabular}
\end{minipage}
\vspace{-4mm}
\end{table*}

\subsection{Entities and Relationships}\label{sect:Graph_Structure}

Following manual inspection of the CTINexus reports and consultation with cybersecurity domain experts, we restricted the graph schema to 17 analyst-relevant entity types: ThreatActor, Malware, Tool, Victim, C2\_Infrastructure, Campaign, Incident, Date, Sector, Region, Country, Technique, CVE, Motivation, Mitigation, Capability, and Source. The domain experts validated that the selected entity types are relevant for practical CTI analysis and reflect concepts commonly used by analysts. These entity types correspond to commonly used concepts in CTI ontologies and knowledge graphs, which typically model key elements such as threat actors, malware, vulnerabilities, techniques, and targets to represent cyber attack behavior~\cite{OpenCyKG2021, KnowCTI2024}.
We further restricted the ontology to 20 relationship types capturing frequently occurring interaction patterns in CTI reporting: attacked, uses, exploits, abuses, targets, includes, occurred\_on, has\_alias, attributed\_to, involved\_malware, involved\_tool, used\_technique, occurred\_in, targets\_sector, located\_in, motivated\_by, exploited\_in, mitigates, leverages, and supported\_by. Such schema-based representations of entities 
and relationships are widely used in threat intelligence knowledge graph construction to structure complex CTI data before downstream analysis and reasoning \cite{OpenCyKG2021, KnowCTI2024}. The selected relationships reflect common operational semantics such as attack execution, exploitation, attribution, targeting, temporal occurrence, and mitigation. The schema was validated by cybersecurity analysts and is not 
intended to be exhaustive, but rather to provide a controlled and reproducible ontology for evaluating graph-based retrieval. All entities contain property fields such as name and summary, while relationships include additional attributes, like a date property that enables a temporal dimension for graph triplets. Entities are connected via relationships to form subject-predicate-object triplets, which serve as the fundamental building blocks of the graph.

To transform unstructured CTI text into graph representations, we instruct a GPT-5.2 (gpt-5.2-2025-12-11)\footnote{\url{https://platform.openai.com/docs/models/gpt-5.2}} model using a structured prompt\footnote{\url{https://github.com/ait-cti/beyond-vanilla-rag/blob/main/text-to-cypher-prompt.md}} that specifies the allowed entities, relationships, and valid triplet patterns. This model was selected due to strong reported performance on text-to-code tasks, particularly when applied in an agentic generation setting~\cite{FAIRCodeGen2025}.


\subsection{Schema Validation Analysis}
Schema validation analysis (Tables~\ref{tab:schema_validation} and~\ref{tab:entity_breakdown}) confirms that the generated graph instances strictly conform to the predefined ontology. Every entity node uses one of the 17 permitted types defined in the schema (1{,}044/1{,}044), and every relationship uses one of the 20 permitted edge labels (438/438). All 30 country-code properties adhere to the ISO~3166-1 alpha-2 standard. CVE extraction recall is computed by extracting all CVE identifiers from the source reports via regular-expression matching of the canonical \texttt{CVE-YYYY-NNNNN} pattern and measuring the proportion of these ground-truth mentions that are instantiated as CVE nodes in the graph. 
Across all runs, all 63 CVE identifiers present in the source reports were successfully extracted (100\%). No analogous recall calculation is possible for other entity types such as malware or threat actors, because these lack a standardised identifier format that would allow reliable automated matching between source text and extracted
nodes. Table~\ref{tab:entity_breakdown} therefore reports raw node counts by entity type. In total, GPT-5.2 extracted 101 unique threat actors, 73 unique malware families, 50 unique CVEs, and 47 unique tools across the 10 runs. Evidence attribution, an \texttt{evidence} property containing a supporting quote from the source report, is present on 40.6\% of relationships (178/438). Finer-grained provenance in the form of \texttt{source\_id} and \texttt{page} properties appears on 23.3\% and 20.3\% of relationships, respectively.

\begin{table*}[t]
\centering

\begin{minipage}[t]{0.42\linewidth}
\centering
\caption{Extracted Entity and Relationship Counts
(aggregated over 10 runs; ``Unique'' counts deduplicated by
node identifier across runs).}
\label{tab:entity_breakdown}
\begin{tabular}{lcc}
\toprule
\textbf{Type} & \textbf{Total nodes} & \textbf{Unique} \\
\midrule
\multicolumn{3}{l}{\textbf{Entity Types}} \\
ThreatActor        & 158 & 101 \\
Technique          & 141 & --  \\
Malware            & 109 & 73  \\
Incident           & 102 & 99  \\
Victim             & 88  & 64  \\
CVE                & 67  & 50  \\
Tool               & 67  & 47  \\
Sector             & 62  & 11  \\
Date               & 58  & --  \\
Source             & 52  & 48  \\
Motivation         & 36  & --  \\
Country            & 30  & 12  \\
Capability         & 23  & 15  \\
C2\_Infrastructure & 16  & --  \\
Campaign           & 15  & 15  \\
Region             & 10  & 5   \\
Mitigation         & 10  & 9   \\
\midrule
\textbf{Total}     & \textbf{1{,}044} & -- \\
\bottomrule
\end{tabular}
\end{minipage}
\hfill
\begin{minipage}[t]{0.55\linewidth}
\centering
\caption{The chosen models for the evaluation}
\label{tab:freq}
\begin{tabular}{ccl}
\toprule
\textbf{LLM} & \textbf{\# Parameter} & \textbf{Comment} \\
\midrule
GPT-5.2           & Unknown & Reasoning \\
GPT-4.1-mini      & Unknown & Common in math \\
Kimi K2 Thinking  & 1T      & Largest OS LLM \\
Mistral Small     & 24B     & Instruct \\
Mixtral 8x7B      & 56B     & Instruct \\
\bottomrule
\end{tabular}
\end{minipage}
\vspace{-5mm}
\end{table*}

\subsection{LLM Choice}
The requirements for selecting the baseline LLM are as follows:
\begin{enumerate}
    \item \textbf{Strong long-context performance}, required to reliably process and reason over the large Neo4j database schema.
    \item \textbf{Strong code-generation capability}, necessary for accurately translating natural-language prompts into Cypher queries.
\end{enumerate}

The choice of LLM for prompt-to-Cypher translation was informed by results reported on the SWE-Bench Pro benchmark\footnote{\url{https://vertu.com/ai-tools/gpt-5-2-benchmark-analysis-performance-comparison-vs-gpt-5-1-gemini-3-pro}} and long-context performance benchmarks such as Tau2-Bench-Telecom\footnote{\url{https://artificialanalysis.ai/evaluations/tau2-bench}}~\cite{barres2025tau2benchevaluatingconversationalagents}. At the time of writing, OpenAI’s GPT-5.2 demonstrates leading performance on both benchmarks, outperforming other proprietary models such as Gemini 3 Pro and Claude Opus 4.5. In addition, GPT-5.2 achieves state-of-the-art performance on agentic text-to-code tasks~\cite{FAIRCodeGen2025}, making it well suited for the complex prompt-to-Cypher translation and iterative reasoning required by our graph-based RAG systems.

We selected \emph{Kimi K2 Thinking} as the largest open-source model according to Hugging Face’s model listing\footnote{\url{https://huggingface.co/models?pipeline_tag=text-generation&sort=most_params}}, and used it as a direct open-source competitor to GPT-5.2. Both models are reasoning-oriented, which makes a comparison meaningful, even though GPT-5.2 is probably larger in terms of parameter count than its open-source counterpart. 

Because reasoning models can increase end-to-end latency-especially for GRAG, AGRAG, and HRAG, which involve iterative query generation and execution-we also include smaller, non-reasoning models. In particular, we evaluate GPT-4.1-mini\footnote{\url{https://platform.openai.com/docs/models/gpt-4.1-mini}} and Mistral Small 24B Instruct\footnote{\url{https://mistral.ai/news/mistral-small-3/}}, which is reported to outperform GPT-4.0-mini and is expected to be competitive with GPT-4.1-mini. Finally, to study performance with a medium-sized open-source model in the 50-70B parameter range, we include Mixtral 8x7B Instruct, which outperforms Llama 2 70B on many benchmarks\footnote{\url{https://huggingface.co/mistralai/Mixtral-8x7B-Instruct-v0.1}}.

Together, this model-set (see Table~\ref{tab:freq}) is intentionally heterogeneous: it includes a strong long-context reasoning models to approximate an upper bound for structured tool use, smaller proprietary and open-source baselines representative of cost-sensitive deployments, and large open-source reasoning models to test whether graph-based RAG gains generalize beyond a single vendor. This diversity allows us to attribute observed performance differences primarily to retrieval architecture and agentic design choices, rather than to a specific model family or scale.

\subsection{Evaluation Metrics}
In this study, we employ two complementary sets of evaluation metrics.

\begin{itemize}
    \item \textbf{Classical automatic metrics}, including F1-Score~\cite{vanrijsbergen1979information}, BLEU~\cite{blagec-etal-2022-global}, ROUGE-1~\cite{lin-2004-rouge} and ROUGE-L~\cite{lin-2004-rouge}, and BERTScore (F1)~\cite{zhang2020bertscoreevaluatingtextgeneration}.
    \item \textbf{LLM-as-a-Judge evaluation}, using a custom-defined rating schema.
\end{itemize}

The motivation for using both metric types is that classical metrics can be overly sensitive to surface-level differences in wording or formatting. For example, a ground-truth answer such as ``First of January, 2020'' may be penalized when the model produces ``01.01.2020'', despite the two answers being semantically equivalent. Classical metrics often struggle to robustly capture this type of semantic equivalence~\cite{mehra2025improvingapplicabilitydeeplearning}. Consequently, LLM-based evaluation is included to better assess semantic correctness and answer quality beyond lexical overlap~\cite{zheng2023judging,liu2023geval,Gu2025}.

\paragraph{Composite classical metric.}
In addition to reporting individual classical metrics, we aggregate F1, BLEU, ROUGE-1, ROUGE-L, and BERTScore (F1) into a single composite score to facilitate concise comparison across systems.
Each metric is aggregated using equal weights. BLEU is used in its native scale, while the remaining metrics naturally lie in the $[0,1]$ range; the composite score is intended as a comparative summary rather than a calibrated absolute measure.

This yields the following weighted average:

\begin{equation}
\text{CompositeScore} =
\frac{1}{5} \bigl(
\text{F1}
+ \text{BLEU}
+ \text{ROUGE-1}
+ \text{ROUGE-L}
+ \text{BERTScore}_{\mathrm{F1}}
\bigr)
\end{equation}


Equal weighting avoids privileging any single surface-form metric and reflects our intent to treat lexical overlap and semantic similarity as equally informative signals.

\subsection{LLM-as-a-Judge Evaluation}
\label{subsec:llm_judge}

To mitigate the limitations of surface-form–based metrics, we employ an \emph{LLM-as-a-Judge} evaluation framework, in which a large language model compares candidate answers against a reference (baseline) answer using a fixed scoring rubric. Each candidate is evaluated along four criteria, scored on a 0-5 scale and combined via weighted aggregation:

\begin{itemize}
    \item \textbf{C1: Agreement with Baseline (weight 4).} Measures semantic alignment with the baseline answer, independent of surface form. Contradictions and unsupported deviations are penalized.
    \item \textbf{C2: Task Adequacy (weight 3).} Assesses how completely and correctly the candidate addresses the question. Explicitly acknowledging insufficient information is considered fully adequate when the context does not support an answer.
    \item \textbf{C3: Faithfulness (weight 2).} Evaluates whether the answer avoids hallucinations and unsupported claims, remaining grounded in the available context.
    \item \textbf{C4: Clarity and Brevity (weight 1).} Rewards clear, concise, and well-structured responses.
\end{itemize}

The final score, or answer quality, is computed as:
\[
\text{Weighted Total}= \text{Answer Quality} = 4C1 + 3C2 + 2C3 + C4,
\]
with a maximum of 50 points.

If the baseline answer is empty or explicitly states that the question cannot be answered, candidates that clearly acknowledge the lack of sufficient information-without adding speculative content-receive full scores for agreement, adequacy, and faithfulness. This prevents penalizing correct refusals in under-specified settings.

The judge produces a strictly structured JSON object\footnote{\url{https://github.com/ait-cti/beyond-vanilla-rag/blob/main/llm-judge.md}} containing per-system criterion scores, weighted totals, short qualitative comments, and an explicit ranking. Prior work shows that LLM-based evaluators correlate strongly with human judgments and outperform classical automatic metrics in assessing semantic correctness and factual consistency, particularly for open-ended generation tasks \cite{zheng2023judging, liu2023geval, Gu2025, fu2023gptscore}.

\begin{table*}[t]
\centering

\begin{minipage}{0.60\linewidth}
\centering
\caption{Inter-criterion Pearson correlations (N=2{,}640 judgments across 10 runs). 
C1--C3 form a tightly coupled correctness cluster ($r \geq 0.91$), 
while C4 (Clarity) is substantially less correlated ($r = 0.43$--$0.50$).}
\label{tab:criterion_correlations}
\begin{tabular}{lcccc}
\toprule
 & \textbf{C1} & \textbf{C2} & \textbf{C3} & \textbf{C4} \\
 & Agreement & Adequacy & Faithfulness & Clarity \\
\midrule
C1 (Agreement) & --- & \textbf{0.98} & \textbf{0.92} & 0.43 \\
C2 (Adequacy) &  & --- & \textbf{0.91} & 0.46 \\
C3 (Faithfulness) &  &  & --- & 0.50 \\
C4 (Clarity) &  &  &  & --- \\
\bottomrule
\end{tabular}
\end{minipage}
\hfill
\begin{minipage}{0.36\linewidth}
\centering
\caption{Per-run stability (mean $\pm$ sd across 10 runs)}
\label{tab:per_run_stability}
\begin{tabular}{lc}
\toprule
\textbf{Pair} & \textbf{$r$} \\
\midrule
C1–C2 & $0.98 \pm 0.01$ \\
C1–C3 & $0.92 \pm 0.02$ \\
C2–C3 & $0.91 \pm 0.02$ \\
C1–C4 & $0.42 \pm 0.07$ \\
C2–C4 & $0.45 \pm 0.07$ \\
C3–C4 & $0.49 \pm 0.06$ \\
\bottomrule
\end{tabular}
\end{minipage}

\end{table*}

\paragraph{Rubric validation.}
Post-hoc analysis (see Table~\ref{tab:criterion_correlations} and Table~\ref{tab:per_run_stability}) of inter-criterion correlations supports the design of the scoring rubric. C1 (Agreement), C2 (Adequacy), and C3 (Faithfulness) are highly correlated ($r{=}0.92$ to $0.98$), reflecting a shared underlying correctness construct: answers that agree with the baseline also tend to be adequate and faithful. In contrast, C4 (Clarity) is substantially less correlated with the other criteria ($r{=}0.43$ to $0.50$), confirming that it captures an orthogonal quality dimension. This supports our answer quality metric design, which assigns the highest weight to Agreement (C1, weight~4) and the lowest weight to Clarity (C4, weight~1), ensuring the composite score is dominated by factual correctness rather than surface-form quality. Additionally, the near-independence of C4 provides a diagnostic signal: systems that score high on C4 but low on C1--C3 produce fluent but incorrect outputs, a hallmark of confident hallucination.

\section{Results and Analysis}
\label{sec:results}

This section presents the results and answers the research questions. Additionally, a detailed adversary model failure analysis is performed and elaborated.

\subsection{Answers to Research Questions}

\subsubsection{Semantic RAG vs Others on CTI tasks.}

We quantify improvements using LLM-as-a-Judge assessment and classical automatic metrics (F1, BLEU, ROUGE, BERTScore). Table~\ref{tab:rq1_judge_improvements} shows AGRAG and HRAG deliver large improvements over RAG ($+10.68$ and $+11.68$ points respectively) with medium effect sizes (Cohen's $d{=}0.43$ and $0.52$). GRAG provides only marginal benefits ($+1.34$, zero median), suggesting schema-constrained query errors offset graph grounding advantages without correction mechanisms. Table~\ref{tab:rq1_classic_improvements} corroborates these findings: AGRAG and HRAG exceed twofold improvements on composite classical metrics with large effect sizes, while GRAG slightly underperforms RAG.

\begin{tcolorbox}[
  colback=gray!10,
  colframe=black,
  boxrule=0.8pt,
  arc=4pt,
  left=6pt,
  right=6pt,
  top=6pt,
  bottom=6pt
]
\textbf{RQ1: To what extent do graph-based and hybrid RAG systems improve answer quality over semantic RAG on CTI tasks?} Graph-based systems substantially improve CTI answer quality only when augmented with agentic correction or hybrid retrieval.
\end{tcolorbox}

\begin{table}[t]
\centering
\caption{LLM-as-a-Judge improvements over RAG (RQ1). Mean and median differences are computed over 3{,}300 paired question instances.}
\label{tab:rq1_judge_improvements}
\begin{tabular}{lcccc}
\toprule
\textbf{System} & \textbf{Mean $\Delta$ vs RAG} & \textbf{95\% Bootstrap CI} & \textbf{Median $\Delta$} & \textbf{Cohen's $d$} \\
\midrule
GRAG  & $+1.34$  & $[0.40,\, 2.28]$   & $0$  & $0.05$ \\
AGRAG & $+10.68$ & $[9.84,\, 11.52]$  & $+4$ & $0.43$ \\
HRAG  & $+11.68$ & $[10.88,\, 12.45]$ & $+4$ & $0.52$ \\
\bottomrule
\end{tabular}
\vspace{-4mm}
\end{table}

\begin{table}[t]
\centering
\caption{Classical-metric improvements over vanilla RAG (RQ1). Mean and median differences are computed over 3{,}300 paired question instances using the equally weighted composite score of F1, BLEU, ROUGE-1, ROUGE-L, and BERTScore.}
\label{tab:rq1_classic_improvements}
\begin{tabular}{lcccc}
\toprule
\textbf{System} & \textbf{Mean $\Delta$ vs RAG} & \textbf{95\% Bootstrap CI} & \textbf{Median $\Delta$} & \textbf{Cohen's $d$} \\
\midrule
GRAG  & $-0.06$ & $[-0.53,\, 0.37]$ & $+0.12$ & $-0.05$ \\
AGRAG & $+1.93$ & $[0.83,\, 3.01]$  & $+2.19$ & $0.76$ \\
HRAG  & $+1.80$ & $[0.92,\, 2.68]$  & $+1.68$ & $0.88$ \\
\bottomrule
\end{tabular}
\vspace{-5mm}
\end{table}


\begin{figure}
    \centering
    \includegraphics[width=0.6\linewidth]{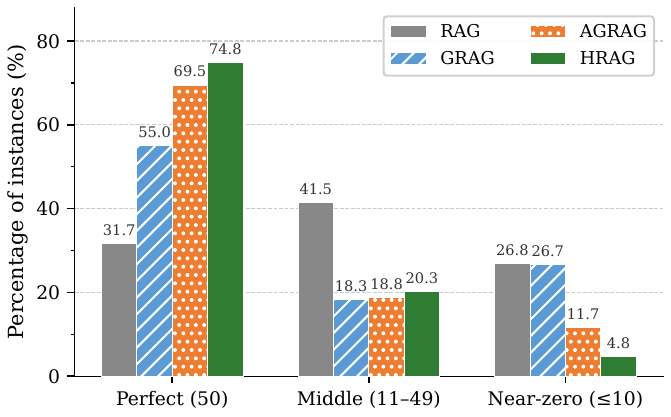}
    \caption{Score distribution across RAG systems (LLM-as-a-Judge, 0--50 scale). Each cell reports the percentage of the evaluated question instances falling in the indicated score range}
    \label{fig:score_distribution}
    \vspace{-8mm}
\end{figure}

\paragraph{Score Distribution Analysis} Mean improvements alone obscure a critical distributional phenomenon. Fig.~\ref{fig:score_distribution} reveals that all graph-based systems exhibit \emph{bimodal} score distributions, answers cluster at either perfect (50) or near-failure ($\leq$10) scores, with relatively few intermediate outcomes.

GRAG shows the most extreme bimodality: while 55.0\% of answers achieve a perfect score (higher than RAG's 31.7\%), a nearly identical fraction of answers collapse to near-zero (26.7\%) as in RAG (26.8\%). Only 18.3\% of GRAG answers fall in the intermediate range, compared to 41.5\% for RAG. This ``all-or-nothing'' pattern explains GRAG's marginal mean improvement (+1.34) despite producing more perfect answers than RAG: the gains from successful graph queries are offset by catastrophic failures when Cypher generation or schema matching fails.

In contrast, AGRAG and HRAG progressively reduce the failure tail. AGRAG cuts near-zero answers from 26.7\% (GRAG) to 11.7\%, while HRAG reduces them further to 4.8\%, a 5.6-fold reduction relative to GRAG. Simultaneously, HRAG achieves the highest perfect-score rate (74.8\%). This demonstrates that agentic correction and hybrid redundancy do not merely improve average performance; they fundamentally reshape the failure distribution, converting catastrophic collapses into either correct answers or graceful degradations.

\begin{tcolorbox}[
  colback=gray!10,
  colframe=black,
  boxrule=0.8pt,
  arc=4pt,
  left=6pt,
  right=6pt,
  top=6pt,
  bottom=6pt
]
\textbf{Summary.} Graph-based systems exhibit strongly bimodal performance, producing either perfect answers or near-failures, with GRAG showing an extreme ``all-or-nothing'' pattern that offsets its gains through frequent collapses. In contrast, AGRAG and especially HRAG significantly reduce these failures and increase perfect answers, demonstrating that agentic correction and hybrid retrieval reshape the distribution toward more reliable outcomes rather than just improving average performance.
\end{tcolorbox}

\subsubsection{Benefits of Graph Grounding on Different Question Types}

\begin{table}[t]
\centering
\caption{RQ2 (LLM-as-a-Judge): Improvements over semantic RAG by question type. Mean differences are shown with 95\% bootstrap confidence intervals. $n$ denotes the number of paired question instances.}
\label{tab:rq2_judge_by_category}
\begin{tabular}{llccccc}
\toprule
\textbf{Category} & \textbf{System} & \textbf{Mean $\Delta$ vs RAG} & \textbf{95\% CI} & \textbf{Median $\Delta$} & \textbf{Cohen's $d$} \\
\midrule
guided & AGRAG  & $-8.74$  & $[-10.29,\, -7.19]$ & $-7.0$  & $-0.39$ \\
guided & GRAG   & $-17.67$ & $[-19.17,\, -16.13]$ & $-18.0$ & $-0.81$ \\
guided & HRAG   & $+4.32$  & $[2.74,\, 5.86]$ & $+1.0$  & $0.19$ \\
\midrule
multi\_hop & AGRAG & $+17.64$ & $[16.04,\, 19.29]$ & $+26.0$ & $0.78$ \\
multi\_hop & GRAG  & $+10.37$ & $[8.61,\, 12.17]$ & $+4.0$  & $0.41$ \\
multi\_hop & HRAG  & $+15.18$ & $[13.69,\, 16.69]$ & $+9.5$  & $0.72$ \\
\midrule
simple & AGRAG & $+22.81$ & $[21.33,\, 24.31]$ & $+33.0$ & $1.11$ \\
simple & GRAG  & $+13.04$ & $[11.19,\, 14.85]$ & $+5.0$  & $0.51$ \\
simple & HRAG  & $+18.22$ & $[16.67,\, 19.76]$ & $+18.0$ & $0.85$ \\
\midrule
single\_hop & AGRAG & $+15.42$ & $[13.83,\, 17.01]$ & $+11.0$ & $0.69$ \\
single\_hop & GRAG  & $+5.26$  & $[3.41,\, 7.13]$ & $0.0$   & $0.19$ \\
single\_hop & HRAG  & $+14.98$ & $[13.56,\, 16.45]$ & $+5.0$  & $0.72$ \\
\midrule
unanswerable & AGRAG & $+1.20$  & $[-0.45,\, 2.84]$ & $0.0$   & $0.09$ \\
unanswerable & GRAG  & $-11.80$ & $[-14.37,\, -9.28]$ & $-1.0$  & $-0.56$ \\
unanswerable & HRAG  & $-4.84$  & $[-7.27,\, -2.54]$ & $0.0$   & $-0.25$ \\
\bottomrule
\end{tabular}
\vspace{-3mm}
\end{table}

\begin{table}[t]
\centering
\caption{RQ2 (Classical metrics): Improvements over semantic RAG by question type using the composite score of F1, BLEU, ROUGE-1, ROUGE-L, and BERTScore. Mean differences are shown with 95\% bootstrap confidence intervals over paired (model, category) cells ($n{=}5$ models).}
\label{tab:rq2_classic_by_category}
\begin{tabular}{llcccc}
\toprule
\textbf{Category} & \textbf{System} & \textbf{Mean $\Delta$ vs RAG} & \textbf{95\% CI} & \textbf{Median $\Delta$} & \textbf{Cohen's $d$} \\
\midrule
guided & AGRAG & $-1.74$ & $[-2.44,\, -0.95]$ & $-2.26$ & $-1.81$ \\
guided & GRAG  & $-1.80$ & $[-2.33,\, -1.00]$ & $-2.02$ & $-1.96$ \\
guided & HRAG  & $-0.84$ & $[-1.66,\, -0.25]$ & $-0.49$ & $-0.90$ \\
\midrule
multi\_hop & AGRAG & $+2.10$ & $[1.85,\, 2.31]$ & $+2.07$ & $6.97$ \\
multi\_hop & GRAG  & $+0.44$ & $[0.16,\, 0.78]$ & $+0.38$ & $1.08$ \\
multi\_hop & HRAG  & $+2.09$ & $[1.45,\, 2.68]$ & $+2.57$ & $2.62$ \\
\midrule
simple & AGRAG  & $+4.04$ & $[3.54,\, 4.52]$ & $+3.98$ & $6.63$ \\
simple & GRAG   & $+0.50$ & $[0.19,\, 0.74]$ & $+0.50$ & $1.37$ \\
simple & HRAG   & $+3.30$ & $[2.31,\, 4.29]$ & $+3.28$ & $2.54$ \\
\midrule
single\_hop & AGRAG & $+2.48$ & $[2.01,\, 3.30]$ & $+2.15$ & $2.76$ \\
single\_hop & GRAG  & $+0.34$ & $[0.04,\, 0.79]$ & $+0.13$ & $0.68$ \\
single\_hop & HRAG  & $+2.17$ & $[1.26,\, 3.31]$ & $+2.09$ & $1.61$ \\
\bottomrule
\end{tabular}
\vspace{-4mm}
\end{table}

Table~\ref{tab:rq2_judge_by_category} reveals strong interaction between question type and graph grounding effectiveness. For simple questions, AGRAG achieves $+22.81$ points ($d{=}1.11$), HRAG $+18.22$ ($d{=}0.85$), and GRAG $+13.04$ ($d{=}0.51$), indicating direct fact lookup benefits from graph structure. Single-hop questions show similar patterns: AGRAG $+15.42$, HRAG $+14.98$, GRAG $+5.26$. All graph systems strongly outperform vanilla RAG on multi-hop questions (AGRAG $+17.64$, HRAG $+15.18$, GRAG $+10.37$), confirming explicit grounding aids compositional reasoning where semantic retrieval fails to retrieve distributed evidence.

For guided questions, GRAG ($-17.67$) and AGRAG ($-8.74$) underperform RAG, while HRAG achieves $+4.32$, indicating these questions require both precise facts and contextual synthesis. Pure graph pipelines produce rigid answers; HRAG combines structured and unstructured evidence. For unanswerable questions, GRAG ($-11.80$) and HRAG ($-4.84$) degrade, while AGRAG shows slight improvement ($+1.20$).

\begin{tcolorbox}[
  colback=gray!10,
  colframe=black,
  boxrule=0.8pt,
  arc=4pt,
  left=6pt,
  right=6pt,
  top=6pt,
  bottom=6pt
]
\textbf{RQ2: To what extent do different question types benefit from explicit graph grounding?} Graph grounding significantly improves performance on simple, single-hop, and especially multi-hop questions, where explicit relational structure enables effective fact lookup and compositional reasoning beyond semantic retrieval. However, its effectiveness depends on question type: pure graph approaches underperform on guided and unanswerable queries, while hybrid retrieval (HRAG) achieves the most robust performance by combining structured reasoning with contextual evidence.

\end{tcolorbox}
\vspace{-6mm}


\begin{table*}[t]
\centering

\begin{minipage}[t]{0.45\linewidth}
\centering
\caption{Hallucination rate by system and question category, defined as the percentage of answers receiving a Faithfulness score C3~$\leq$~2 on the 0--5 scale. Lower is better.}
\label{tab:hallucination_rates}
\begin{tabular}{lrrrr}
\toprule
\textbf{Category} & \textbf{RAG} & \textbf{GRAG} & \textbf{AGRAG} & \textbf{HRAG} \\
\midrule
Simple       & 54.7\% & 12.7\% & 12.7\% &  8.0\% \\
Single-hop   & 42.0\% & 13.3\% &  7.3\% &  2.0\% \\
Multi-hop    & 56.7\% & 18.7\% & 20.0\% & 21.3\% \\
Guided       & 41.9\% & \textbf{84.4\%} & 33.1\% & 15.6\% \\
Unanswerable & 10.0\% & 46.0\% &  4.0\% & 20.0\% \\
\midrule
\textbf{Overall} & \textbf{45.8\%} & \textbf{34.1\%} & \textbf{17.4\%} & \textbf{12.4\%} \\
\bottomrule
\end{tabular}
\end{minipage}
\hfill
\begin{minipage}[t]{0.50\linewidth}
\centering
\begin{threeparttable}
\caption{Safety on Unanswerable Questions: System Abstention Behavior}
\label{tab:unanswerable_safety}

\begin{tabular}{lcc}
\toprule
System 
& \makecell{Correct Refusal (\%) \\ (higher better)} 
& \makecell{Overconfident Answer (\%) \\ (lower better)} \\
\midrule
RAG   & 0.0\%  & 100.0\%$^{\dagger}$ \\
GRAG  & 44.0\% & 56.0\%$^{\dagger}$ \\
AGRAG & 0.0\%  & 100.0\%$^{\dagger}$ \\
HRAG  & 76.0\% & 24.0\% \\
\bottomrule
\end{tabular}

\begin{tablenotes}
\item[$\dagger$] \footnotesize Safety-critical failure: system provides confident answers despite insufficient evidence.
\end{tablenotes}

\end{threeparttable}
\end{minipage}
\vspace{-6mm}
\end{table*}

\paragraph{Faithfulness Analysis: Hallucination Rates by Category} To complement the aggregate quality analysis, we examine Faithfulness (C3) scores as a proxy for hallucination behavior. Table~\ref{tab:hallucination_rates} reports the percentage of answers receiving C3~$\leq$~2, indicating unfaithful or unsupported content.

Semantic RAG exhibits the highest overall hallucination rate (45.8\%), with rates exceeding 50\% on simple and multi-hop questions, categories where semantic retrieval frequently fails to surface the specific evidence needed. All graph-based systems substantially reduce hallucination on factual categories: for simple questions, all three graph systems achieve $\leq$12.7\% versus RAG's 54.7\%; for single-hop, HRAG reaches 2.0\%.

However, the category-level breakdown reveals a critical inversion for guided questions: GRAG's hallucination rate (84.4\%; defined as Faithfulness score C3 $\leq 2$ on the 0–5 judge scale) is more than double that of RAG (41.9\%). When the graph schema does not cover the concepts referenced by analyst-style synthesis questions, GRAG produces unfaithful answers grounded in incorrect or empty query results rather than acknowledging uncertainty. AGRAG partially mitigates this (33.1\%) through critique-based detection of invalid queries, while HRAG achieves the lowest guided hallucination rate (15.6\%) by falling back to unstructured evidence when structured paths fail. For unanswerable questions, the pattern differs: RAG achieves a hallucination rate of 10.0\% because its generated answers, while overconfident, tend to align superficially with plausible refusals. GRAG (46.0\%) produces unfaithful outputs from failed query loops, while AGRAG achieves the best faithfulness (4.0\%) on this category.

These findings demonstrate that hallucination risk is not uniformly reduced by graph grounding; rather, it shifts across question types. Graph structure eliminates a large class of factual hallucinations (e.g., incorrect entity 
or relationship claims) but introduces a distinct failure mode when queries fall outside schema coverage. In such cases, the system may generate answers that are formally consistent with the graph output but semantically unsupported by the underlying reports, a phenomenon we refer to as structural hallucination.

\begin{tcolorbox}[
  colback=gray!10,
  colframe=black,
  boxrule=0.8pt,
  arc=4pt,
  left=6pt,
  right=6pt,
  top=6pt,
  bottom=6pt
]
\textbf{Summary.} Graph-based retrieval substantially reduces hallucinations for factual questions compared to semantic RAG, but this benefit is highly dependent on schema coverage and question type. When queries fall outside the graph schema, graph-based systems, especially GRAG, introduce a new failure mode of ``structural hallucination'', whereas hybrid approaches (HRAG) mitigate this by falling back to unstructured evidence.
\end{tcolorbox}

\paragraph{Operational Safety Analysis: Abstention Behavior} Note on metrics. The judge-based deltas in Table~\ref{tab:rq2_judge_by_category} evaluate answer quality on a 0-50 scale, whereas Table~\ref{tab:unanswerable_safety} reports a separate binary outcome (\emph{explicit refusal} vs.\ \emph{attempted answer}). These measures can diverge: a model may earn non-zero judge scores while still attempting an answer, and a refusal can be judged poorly if it is not clearly justified. Table~\ref{tab:unanswerable_safety} shows HRAG achieves the best correct refusal rate (76\%) through graph-text cross-validation. RAG and AGRAG exhibit zero abstention (100\% overconfidence), a critical safety failure where false positives misdirect investigation. GRAG combines 56\% overconfidence with extreme latency: unanswerable queries average 520 seconds (8.7 minutes), with worst-case 2,348 seconds (39.1 minutes), using a large reasoning LLM, exhausting Cypher correction loops. Table~\ref{tab:rq2_classic_by_category} confirm judge findings for factual questions (AGRAG/HRAG achieve $+4.0$ for simple, $+2.1$ for multi-hop), while confirming guided question degradation reflects surface-form metric limitations on explanatory answers. 

\begin{tcolorbox}[
  colback=gray!10,
  colframe=black,
  boxrule=0.8pt,
  arc=4pt,
  left=6pt,
  right=6pt,
  top=6pt,
  bottom=6pt
]
\textbf{Summary.} Graph grounding effectiveness varies substantially by question type. Simple, single-hop, and multi-hop questions benefit most; guided questions require hybrid retrieval; unanswerable questions expose calibration challenges requiring explicit uncertainty mechanisms.
\end{tcolorbox}

\begin{figure*}
    \centering
    \includegraphics[width=1\linewidth]{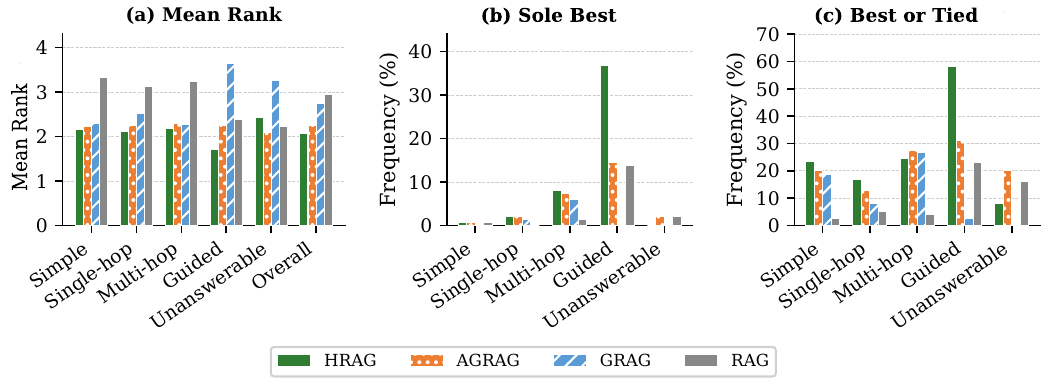}
    \caption{Mean ranks and top-placement frequency by question category (10 runs, LLM-as-a-Judge). (a) Mean rank (1 = best); (b) sole best rate; (c) best-or-tied rate.}
    \label{fig:mean_rank_figure}

    \vspace{-7mm}
\end{figure*}

\paragraph{Rank-based summary.}
To complement score-based analysis, which can be sensitive to outliers, we compute mean ranks (see Fig.~\ref{fig:mean_rank_figure}) across the four systems for each question instance (1\,=\,best, 4\,=\,worst). Overall, HRAG achieves the best mean rank (2.06), followed by AGRAG (2.24), GRAG (2.74), and RAG (2.95). The ranking pattern varies by category: on guided questions, HRAG dominates (rank 1.71, sole best system in 36.9\% of instances), while RAG (2.39) outranks GRAG (3.65)---confirming that unstructured retrieval is preferable to graph-only retrieval for analyst-style synthesis. On unanswerable questions, AGRAG achieves the best rank (2.09), reflecting its low hallucination rate on this category. For factual categories (simple, single-hop, multi-hop), all three graph systems outrank RAG, with HRAG consistently first (see Figure~\ref{fig:mean_rank_figure}).

\subsubsection{Text-to-Cypher Generation}

\begin{table*}[t]
\centering
\caption{RQ3 (LLM impact on text-to-Cypher performance): Judge-based performance of Cypher-dependent systems by underlying LLM. Mean scores and deltas are reported with 95\% bootstrap confidence intervals. Collapse rate indicates the fraction of GRAG answers with near-failure scores ($\leq 10$ 10 on the 0–50 judge scale).}
\label{tab:rq3_llm_cypher}
\begin{tabular}{lcccc}
\toprule
\textbf{Model} &
\textbf{GRAG Mean} &
\textbf{AGRAG$-$GRAG $\Delta$} &
\textbf{HRAG$-$GRAG $\Delta$} &
\textbf{GRAG Collapse Rate} \\
\midrule
Kimi K2 Thinking
& $42.55$ [41.40, 43.65]
& $+1.06$ [0.46, 1.69]
& $+2.65$ [1.87, 3.42]
& $0.04$ \\

Mixtral 8x7B
& $29.91$ [28.52, 31.25]
& $+4.38$ [3.62, 5.12]
& $+9.62$ [8.33, 10.86]
& $0.41$ \\

Mistral Small 24B
& $31.18$ [29.74, 32.55]
& $+3.64$ [2.88, 4.42]
& $+7.88$ [6.59, 9.12]
& $0.36$ \\

GPT-4.1-mini
& $36.49$ [35.36, 37.63]
& $+1.79$ [1.28, 2.30]
& $+3.88$ [3.01, 4.77]
& $0.18$ \\

GPT-5.2
& $33.02$ [31.75, 34.34]
& $+1.64$ [0.96, 2.32]
& $+10.65$ [8.99, 12.35]
& $0.27$ \\
\bottomrule
\end{tabular}
\vspace{-5mm}
\end{table*}

Cypher-dependent retrieval systems (GRAG, AGRAG, and HRAG) rely on an LLM to translate natural-language questions into executable Cypher queries. As a result, their performance depends not only on retrieval architecture, but also on the model’s ability to generate syntactically valid queries, respect the graph schema, and handle empty or invalid query results. Table~\ref{tab:rq3_llm_cypher} shows strong sensitivity of GRAG to the underlying LLM. Kimi K2 Thinking achieves the highest GRAG mean score (42.55) and the lowest collapse rate (4\%), indicating robust text-to-Cypher generation. In contrast, smaller open-source models exhibit substantially higher collapse rates (Mixtral 41\%, Mistral Small 36\%), meaning that a large fraction of GRAG queries degrade into near-failure outputs.

We assess LLM influence using two complementary signals: LLM-as-a-Judge scores (0--50) and a \emph{collapse rate} for GRAG. Collapse is defined as a near-failure output with a judge score $\leq$10/50, typically caused by invalid Cypher, schema mismatches, or prolonged empty-result retry loops. For GPT-5.2, all unanswerable instances collapse (see Table~\ref{tab:collapse_by_category}), whereas collapse remains rare for answerable categories. Unless stated otherwise, collapse-rate breakdowns are reported for GPT-5.2.

Agentic correction (AGRAG) improves GRAG performance across all models, but the magnitude of improvement depends on baseline model strength. Weaker models benefit the most (Mixtral $+4.38$, Mistral Small $+3.64$), while stronger models show only modest gains ($+1$ to $+2$ points). This pattern indicates that AGRAG primarily compensates for syntactic and schema-level errors in Cypher generation rather than fundamentally improving reasoning capability.

Hybrid retrieval (HRAG) yields the most consistent gains across all LLMs. By combining graph-based querying with unstructured evidence retrieval, HRAG can recover useful information even when Cypher generation fails or returns empty results. Consequently, HRAG substantially outperforms GRAG for every model, with particularly large improvements for GPT-5.2 ($+10.65$) and Mixtral ($+9.62$).

\begin{tcolorbox}[
  colback=gray!10,
  colframe=black,
  boxrule=0.8pt,
  arc=4pt,
  left=6pt,
  right=6pt,
  top=6pt,
  bottom=6pt
]
\textbf{RQ3: How does the underlying LLM affect text-to-Cypher generation performance?} Cypher-based retrieval systems depend heavily on the underlying LLM’s ability to generate correct and schema-compliant queries, leading to large performance differences across models and frequent failures for weaker ones. While agentic correction helps reduce these errors, HRAG is the most reliable approach, consistently improving performance by compensating for failed or incomplete graph queries.
\end{tcolorbox}

\begin{table}[t]
\centering
\caption{GRAG collapse rate by question category for GPT-5.2 (collapsed = judge score $\leq 10/50$).}
\label{tab:collapse_by_category}
\begin{tabular}{lcc}
\toprule
\textbf{Category} & \textbf{Collapsed / Total} & \textbf{Collapse Rate} \\
\midrule
Unanswerable & 50 / 50 & 100.0\% \\
Guided & 8 / 160 & 5.0\% \\
Multi-hop & 7 / 150 & 4.7\% \\
Single-hop & 1 / 150 & 0.7\% \\
Simple & 0 / 150 & 0.0\% \\
\bottomrule
\end{tabular}
\vspace{-3mm}
\end{table}

\paragraph{Collapse Rate Analysis by Question Type} Table~\ref{tab:collapse_by_category} localizes GRAG failures by question type for the case of evaluation with GPT-5.2 LLM model. Collapse is universal for unanswerable questions (100\%), revealing a structural limitation of graph-only pipelines: when the correct behavior is to acknowledge missing information, GRAG instead continues generating alternative Cypher queries, entering prolonged retry loops (mean 520\,s, maximum 2{,}348\,s). In contrast, collapse is rare for answerable factual categories, particularly simple and single-hop questions, where valid graph paths usually exist. AGRAG mitigates some of these failures by exiting invalid query loops earlier, while HRAG further reduces their impact through fallback retrieval.

\begin{table*}[t]
\centering
\caption{RQ3 (Classical metrics): Impact of the underlying LLM on Cypher-dependent systems. Scores are based on the composite classical metric. Mean values and deltas are reported with 95\% bootstrap confidence intervals.}
\label{tab:rq3_llm_cypher_classic}
\begin{tabular}{lcccc}
\toprule
\textbf{Model} &
\textbf{GRAG Mean} &
\textbf{AGRAG$-$GRAG $\Delta$} &
\textbf{HRAG$-$GRAG $\Delta$} &
\textbf{HRAG$-$RAG $\Delta$} \\
\midrule
Kimi K2 Thinking
& $0.41$ [0.38, 0.44]
& $+0.06$ [0.03, 0.09]
& $+0.12$ [0.08, 0.16]
& $+0.18$ [0.14, 0.22] \\

Mixtral 8x7B
& $0.19$ [0.16, 0.22]
& $+0.09$ [0.06, 0.12]
& $+0.21$ [0.17, 0.25]
& $+0.26$ [0.21, 0.31] \\

Mistral Small 24B
& $0.22$ [0.19, 0.25]
& $+0.08$ [0.05, 0.11]
& $+0.18$ [0.14, 0.22]
& $+0.23$ [0.18, 0.28] \\

GPT-4.1-mini
& $0.29$ [0.26, 0.32]
& $+0.05$ [0.03, 0.07]
& $+0.09$ [0.06, 0.12]
& $+0.14$ [0.11, 0.17] \\

GPT-5.2
& $0.31$ [0.28, 0.34]
& $+0.04$ [0.02, 0.06]
& $+0.23$ [0.18, 0.28]
& $+0.28$ [0.22, 0.34] \\
\bottomrule
\end{tabular}
\vspace{-5.5mm}
\end{table*}

Table~\ref{tab:rq3_llm_cypher_classic} confirms these findings using surface-form metrics. While absolute values differ due to metric scale, the trends are consistent: stronger LLMs yield higher baseline GRAG performance, and HRAG provides the most stable improvements across all model families. Overall, RQ3 shows that text-to-Cypher generation is a critical bottleneck for graph-based RAG. GRAG performance and failure rates vary widely by underlying LLM; agentic critique primarily benefits weaker models by correcting query-generation errors; and hybrid redundancy (HRAG) is the most reliable strategy for mitigating Cypher-induced failures regardless of model strength.

\begin{tcolorbox}[
  colback=gray!10,
  colframe=black,
  boxrule=0.8pt,
  arc=4pt,
  left=6pt,
  right=6pt,
  top=6pt,
  bottom=6pt
]
\textbf{Summary.} GRAG fails consistently on unanswerable questions, entering long retry loops instead of recognizing missing information, while it performs well on simple factual queries where valid graph paths exist. Overall, text-to-Cypher generation is a key bottleneck, with performance strongly depending on the LLM, while HRAG provides the most stable and reliable results across models.
\end{tcolorbox}

\subsubsection{Runtime Quality Trade-offs}

\begin{table}[t]
\centering
\scriptsize
\caption{Results over 10 runs for all models and RAG systems (mean and standard deviation over 10 runs).}
\label{tab:all_runs_all_models}
\begin{tabular}{llcccccccccccc}
\hline
\textbf{Model} & \textbf{System} &
\textbf{R1} & \textbf{R2} & \textbf{R3} & \textbf{R4} & \textbf{R5} &
\textbf{R6} & \textbf{R7} & \textbf{R8} & \textbf{R9} & \textbf{R10} &
\textbf{Mean} & \textbf{Std} \\
\hline

\multirow{4}{*}{GPT-5.2}
& AGRAG & 6.87 & 10.12 & 9.80 & 9.08 & 12.69 & 10.17 & 10.51 & 9.71 & 7.40 & 8.84 & 9.52 & 1.55 \\
& GRAG  & 67.76 & 137.53 & 143.92 & 66.08 & 178.01 & 150.08 & 100.11 & 106.59 & 179.69 & 145.49 & 127.53 & 38.74 \\
& HRAG  & 37.05 & 54.91 & 43.15 & 42.12 & 52.77 & 46.89 & 53.14 & 41.84 & 47.37 & 50.09 & 46.93 & 5.54 \\
& RAG   & 1.27 & 1.30 & 1.22 & 1.38 & 1.50 & 1.25 & 1.26 & 1.23 & 1.34 & 1.41 & 1.32 & 0.08 \\
\hline

\multirow{4}{*}{Mixtral}
& AGRAG & 5.90 & 5.21 & 6.83 & 6.68 & 6.01 & 6.98 & 6.37 & 8.47 & 7.79 & 5.18 & 6.54 & 1.00 \\
& GRAG  & 26.49 & 23.94 & 23.97 & 7.35 & 15.01 & 25.30 & 15.75 & 10.39 & 1.15 & 24.13 & 17.35 & 8.36 \\
& HRAG  & 8.97 & 18.19 & 10.22 & 16.08 & 9.69 & 12.05 & 11.41 & 10.25 & 9.97 & 10.31 & 11.71 & 2.87 \\
& RAG   & 1.29 & 1.41 & 1.30 & 1.36 & 1.40 & 2.58 & 1.44 & 1.31 & 1.45 & 1.36 & 1.49 & 0.37 \\
\hline

\multirow{4}{*}{Kimi-K2 Thinking}
& AGRAG & 11.85 & 11.28 & 10.33 & 13.76 & 12.07 & 10.41 & 10.62 & 10.37 & 10.00 & 11.14 & 11.18 & 1.08 \\
& GRAG  & 42.95 & 38.38 & 39.74 & 44.23 & 51.92 & 40.57 & 28.08 & 32.35 & 28.46 & 38.57 & 38.52 & 6.97 \\
& HRAG  & 43.41 & 43.08 & 37.52 & 45.06 & 41.91 & 39.46 & 35.55 & 41.45 & 37.14 & 46.81 & 41.14 & 3.47 \\
& RAG   & 3.82 & 3.83 & 4.29 & 5.26 & 4.34 & 3.94 & 3.72 & 3.44 & 3.53 & 4.11 & 4.03 & 0.50 \\
\hline

\multirow{4}{*}{GPT-4.1-mini}
& AGRAG & 4.28 & 4.41 & 4.07 & 4.08 & 4.51 & 4.06 & 4.63 & 4.73 & 4.25 & 4.45 & 4.35 & 0.23 \\
& GRAG  & 5.92 & 6.72 & 5.09 & 6.42 & 5.22 & 5.22 & 5.44 & 7.82 & 5.78 & 6.10 & 5.97 & 0.80 \\
& HRAG  & 4.76 & 4.94 & 4.08 & 4.47 & 4.92 & 4.44 & 4.73 & 4.79 & 4.50 & 4.37 & 4.60 & 0.26 \\
& RAG   & 1.26 & 1.40 & 1.17 & 1.24 & 1.24 & 1.24 & 1.24 & 1.23 & 1.15 & 1.20 & 1.24 & 0.06 \\
\hline

\multirow{4}{*}{Mistral Small 24B}
& AGRAG & 4.35 & 4.35 & 3.85 & 4.06 & 3.73 & 3.94 & 4.69 & 3.95 & 3.72 & 3.67 & 4.03 & 0.32 \\
& GRAG  & 3.02 & 20.54 & 0.83 & 16.37 & 26.89 & 10.62 & 8.77 & 29.77 & 20.53 & 16.23 & 15.36 & 9.08 \\
& HRAG  & 4.76 & 6.63 & 8.42 & 5.50 & 4.33 & 6.14 & 7.29 & 4.81 & 9.08 & 4.87 & 6.18 & 1.56 \\
& RAG   & 0.83 & 0.80 & 0.79 & 1.23 & 0.87 & 0.83 & 0.86 & 0.79 & 0.98 & 0.73 & 0.87 & 0.14 \\
\hline

\end{tabular}
\vspace{-5mm}
\end{table}

\begin{table*}[t]
\centering
\caption{RQ4 Summary: Runtime-quality trade-offs across RAG systems. Ranges are computed from per-model mean and standard deviation runtimes in Table~\ref{tab:all_runs_all_models}. Relative cost is computed per model as the ratio of system mean runtime to the corresponding RAG mean runtime; ranges report the minimum and maximum ratios across evaluated models.}
\label{tab:rq4_runtime_summary}
\begin{tabular}{lccccp{3.4cm}}
\toprule
\textbf{System} &
\textbf{Mean Runtime (s)} &
\textbf{Std (s)} &
\textbf{Cost vs RAG} &
\textbf{Variance} &
\textbf{Trade-off Summary} \\
\midrule
RAG
& $[0.87,\ 4.03]$
& $[0.06,\ 0.50]$
& $\times 1$
& Low
& Fast but weakest performance on complex CTI queries \\

AGRAG
& $[4.03,\ 11.18]$
& $[0.23,\ 1.55]$
& $\times [2.77,\ 7.21]$
& Low--Mod.
& Best balance: large quality gains with stable overhead \\

HRAG
& $[4.60,\ 46.93]$
& $[0.26,\ 5.54]$
& $\times [3.71,\ 35.55]$
& Moderate
& Maximum robustness at higher but predictable cost \\

GRAG
& $[5.97,\ 127.53]$
& $[0.80,\ 38.74]$
& $\times [4.81,\ 96.61]$
& High
& Unfavorable: high cost and instability \\
\bottomrule
\end{tabular}
\vspace{-4mm}
\end{table*}

Table~\ref{tab:rq4_runtime_summary} summarizes runtime-quality trade-offs. Classical RAG achieves sub-4s runtimes with minimal variance but poor performance on complex queries. GRAG exhibits highest cost (>120s) and instability from repeated correction loops without reliable quality gains. AGRAG stabilizes execution (4 to 11s) with large quality improvements, yielding best cost-quality balance. HRAG incurs higher cost (up to 47s) from parallel pipelines but achieves strongest robustness.

\begin{table}[t]
\centering
\small
\caption{Estimated API cost per query (USD) by LLM, retrieval system, and question type. Token budgets are scaled by system complexity (LLM calls: RAG\,=\,1, GRAG\,=\,3, AGRAG\,=\,3.5, HRAG\,=\,5) and category-dependent reasoning effort. Reasoning tokens are included only for reasoning-capable models (GPT-5.2, Kimi K2 Thinking). Costs exclude embedding, infrastructure, and database overhead.}
\label{tab:cost_per_query}
\begin{tabular}{llccccc}
\toprule
\textbf{Model} & \textbf{System} & \textbf{Simple} & \textbf{Single-hop} & \textbf{Multi-hop} & \textbf{Guided} & \textbf{Unanswerable} \\
\midrule

\multirow{4}{*}{GPT-5.2}
 & RAG   & \$0.007 & \$0.008 & \$0.011 & \$0.010 & \$0.008 \\
 & GRAG  & \$0.021 & \$0.028 & \$0.042 & \$0.036 & \$0.025 \\
 & AGRAG & \$0.025 & \$0.031 & \$0.045 & \$0.039 & \$0.029 \\
 & HRAG  & \$0.039 & \$0.050 & \$0.074 & \$0.065 & \$0.046 \\
\midrule

\multirow{4}{*}{GPT-4.1-mini}
 & RAG   & \$0.001 & \$0.001 & \$0.001 & \$0.001 & \$0.001 \\
 & GRAG  & \$0.002 & \$0.002 & \$0.002 & \$0.002 & \$0.002 \\
 & AGRAG & \$0.004 & \$0.004 & \$0.004 & \$0.004 & \$0.004 \\
 & HRAG  & \$0.005 & \$0.005 & \$0.005 & \$0.005 & \$0.005 \\
\midrule

\multirow{4}{*}{Kimi K2 T.}
 & RAG   & \$0.004 & \$0.004 & \$0.005 & \$0.005 & \$0.004 \\
 & GRAG  & \$0.010 & \$0.012 & \$0.016 & \$0.015 & \$0.011 \\
 & AGRAG & \$0.013 & \$0.015 & \$0.019 & \$0.017 & \$0.014 \\
 & HRAG  & \$0.020 & \$0.023 & \$0.030 & \$0.027 & \$0.022 \\
\midrule

\multirow{4}{*}{Mistral S. 24B}
 & RAG   & \$0.0003 & \$0.0003 & \$0.0003 & \$0.0003 & \$0.0003 \\
 & GRAG  & \$0.0006 & \$0.0006 & \$0.0006 & \$0.0006 & \$0.0006 \\
 & AGRAG & \$0.0009 & \$0.0009 & \$0.0009 & \$0.0009 & \$0.0009 \\
 & HRAG  & \$0.0013 & \$0.0013 & \$0.0013 & \$0.0013 & \$0.0013 \\
\midrule

\multirow{4}{*}{Mixtral 8x7B}
 & RAG   & \$0.002 & \$0.002 & \$0.002 & \$0.002 & \$0.002 \\
 & GRAG  & \$0.004 & \$0.004 & \$0.004 & \$0.004 & \$0.004 \\
 & AGRAG & \$0.005 & \$0.005 & \$0.005 & \$0.005 & \$0.005 \\
 & HRAG  & \$0.008 & \$0.008 & \$0.008 & \$0.008 & \$0.008 \\
\bottomrule

\end{tabular}
\vspace{-5mm}
\end{table}

Table~\ref{tab:cost_per_query} (API pricing (per 1M tokens): GPT-5.2 \$1.75/\$14.00 (input/output), GPT-4.1-mini \$0.40/\$1.60, Kimi K2 Thinking \$1.20/\$4.00 (Together AI), Mistral Small 24B \$0.10/\$0.30 (Together AI), Mixtral 8x7B \$0.60/\$0.60 (Together AI). Prices as of January 2026.) estimates per-query API costs across all five models. Cost is dominated by model choice over retrieval architecture: GPT-5.2 HRAG costs 30 to 57$\times$ more than Mistral Small HRAG. Cost variation across question types is substantial only for reasoning models (GPT-5.2, Kimi K2 Thinking) due to category-dependent reasoning tokens; non-reasoning models show near-uniform per-query costs driven primarily by the number of LLM calls. These estimates exclude GRAG's wasted tokens from failed correction loops, making its effective cost per correct answer substantially higher.

\begin{table}[t]
\centering
\caption{Cross-run quality stability: coefficient of variation (CV) of run-level mean judge scores by system and question category, computed over 10 independent runs. Lower CV indicates more stable performance across different report samples.}
\label{tab:cross_run_stability}
\begin{tabular}{lcccc}
\toprule
\textbf{Category} & \textbf{RAG} & \textbf{GRAG} & \textbf{AGRAG} & \textbf{HRAG} \\
\midrule
Simple       & 13.7\% &  6.5\% &  9.1\% &  \textbf{3.3\%} \\
Single-hop   & 13.2\% & 15.4\% &  6.3\% &  \textbf{1.9\%} \\
Multi-hop    & 24.1\% & 16.3\% & 10.9\% &  \textbf{7.0\%} \\
Guided       & 12.5\% & 26.5\% & 15.9\% &  \textbf{8.7\%} \\
Unanswerable &  9.2\% & \textbf{44.9\%} & 10.9\% & 14.5\% \\
\midrule
\textbf{Overall (std of run means)} & 1.76 & 2.14 & 1.25 & \textbf{0.76} \\
\bottomrule
\end{tabular}
\vspace{-6mm}
\end{table}

\begin{tcolorbox}[
  colback=gray!10,
  colframe=black,
  boxrule=0.8pt,
  arc=4pt,
  left=6pt,
  right=6pt,
  top=6pt,
  bottom=6pt
]
\textbf{RQ4: How large is the runtime-quality trade-off across systems and models?} RAG is fastest but performs poorly on complex queries, while GRAG is slow and unstable without consistent quality gains due to repeated query loops. AGRAG offers the best balance between cost and performance, and HRAG achieves the highest robustness at higher but predictable runtime and cost, which are mainly driven by the choice of LLM rather than the retrieval architecture.
\end{tcolorbox}

\paragraph{Cross-Run Quality Stability} Beyond runtime stability, operational CTI systems require \emph{quality stability} across different report samples. Table~\ref{tab:cross_run_stability} reports the coefficient of variation (CV) of run-level mean judge scores across the 10 independent evaluation runs, each using a different random sample of 15~CTI reports.

CV is calculated as follows: for each system $s$ and category $c$, let $\bar{x}_{s,c,r}$ denote the mean judge score for run $r \in \{1,\dots,10\}$. The coefficient of variation is
\begin{equation}
  \mathrm{CV}_{s,c} 
  = \frac{\sigma\!\bigl(\bar{x}_{s,c,1},\dots,\bar{x}_{s,c,10}\bigr)}%
         {\mu\!\bigl(\bar{x}_{s,c,1},\dots,\bar{x}_{s,c,10}\bigr)}
  \times 100\%
\end{equation}
where $\sigma(\cdot)$ and $\mu(\cdot)$ are the sample standard deviation and 
mean, respectively. Lower CV indicates more stable performance across different 
report samples.

 HRAG achieves the lowest CV across all answerable categories (1.9\% to 8.7\%), indicating that its quality is largely invariant to the specific reports selected. In contrast, GRAG exhibits extreme instability on unanswerable questions (CV\,=\,44.9\%), meaning that its performance on this safety-critical category fluctuates wildly depending on the graph structure produced by each report sample. This instability compounds GRAG's runtime variance (Table~\ref{tab:rq4_runtime_summary}): not only is GRAG slow and unpredictable in latency, but the \emph{quality} of its outputs is also unpredictable across deployments.

AGRAG substantially stabilizes GRAG's quality variance (overall std reduced from 2.14 to 1.25), while HRAG achieves the tightest bounds (overall std\,=\,0.76). For CTI deployments where consistent performance across heterogeneous report collections is essential, HRAG provides the most reliable quality guarantees.

\begin{figure}
    \centering
    \includegraphics[width=0.6\linewidth]{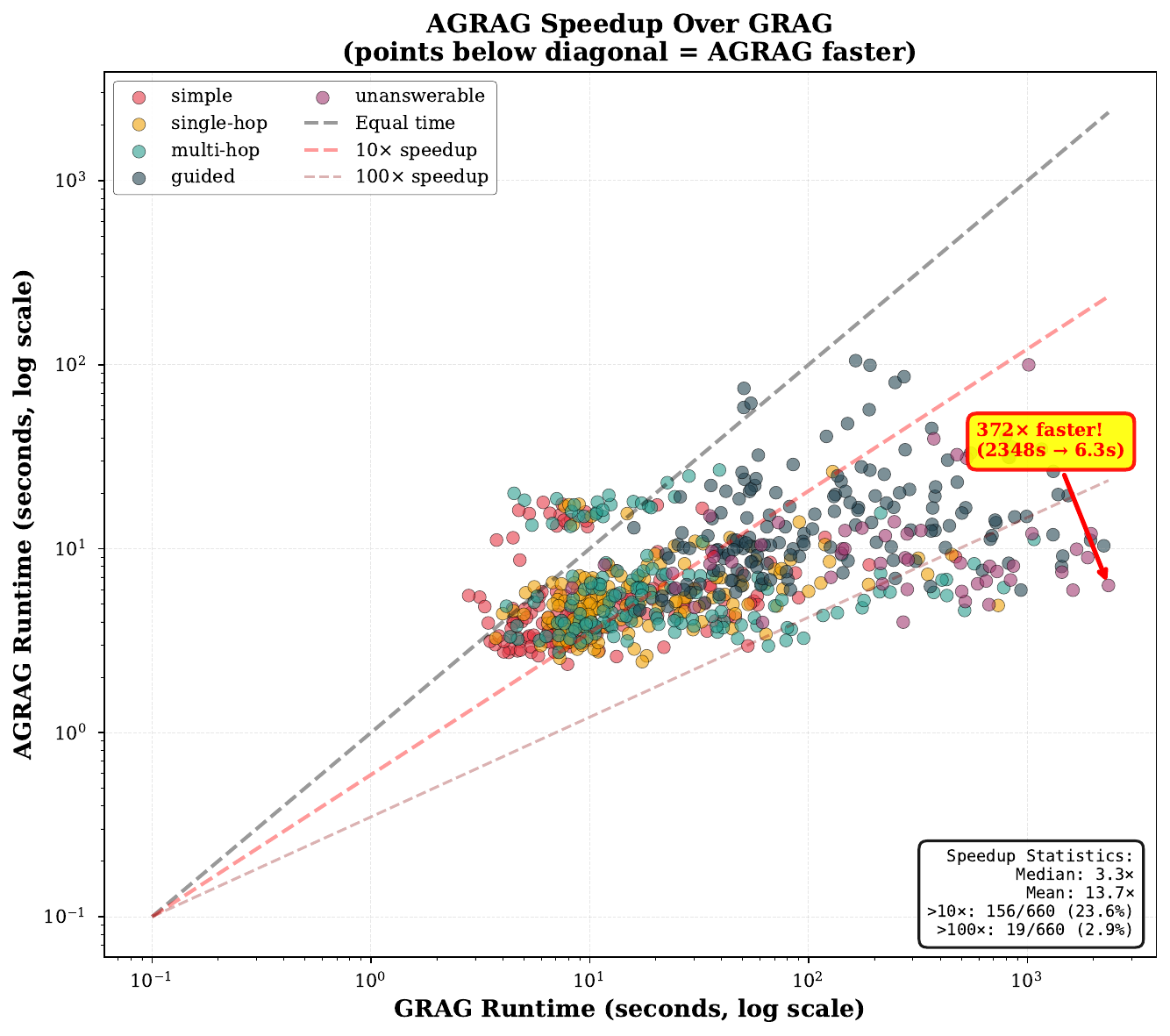}
    \caption{GPT-5.2 Performance Metrics}
    \label{fig:speedup_scatter}
    \caption{AGRAG speedup over GRAG on 660 paired queries. Each point represents one query; points below the diagonal indicate AGRAG faster than GRAG. Median speedup is 3.3×, with 156 queries (23.6\%) exceeding 10× speedup and maximum 372× speedup observed. Purple points (unanswerable queries) cluster in GRAG's extreme timeout region, while AGRAG maintains bounded latency through critique-based early exit.}
    \vspace{-8mm}
\end{figure}

Figure~\ref{fig:speedup_scatter} illustrates GRAG's catastrophic runtime instability versus AGRAG's bounded execution. The scatter plot of 660 paired queries shows that agentic correction not only improves accuracy but can dramatically reduce latency: median speedup is 3.3×, with 23.6\% of queries exceeding 10× speedup and maximum 372× speedup observed. These cases demonstrate that agentic critique identifies invalid queries in the first correction iteration, whereas GRAG's iterative validation becomes trapped attempting variations of fundamentally incorrect query patterns until timeout. Unanswerable queries (purple points) cluster in GRAG's extreme timeout region, while AGRAG maintains predictable sub-60s response times. The trade-off is substantial: RAG minimizes latency but sacrifices quality; GRAG suffers high cost and instability; AGRAG offers best balance for cost-sensitive deployments; HRAG prioritizes robustness over latency. GRAG's extreme variance is intrinsic to naive graph-only designs, representing practical deployment risk in time-sensitive CTI workflows.

\begin{tcolorbox}[
  colback=gray!10,
  colframe=black,
  boxrule=0.8pt,
  arc=4pt,
  left=6pt,
  right=6pt,
  top=6pt,
  bottom=6pt
]
\textbf{Summary.} HRAG achieves the most stable performance across different report samples, while GRAG shows extreme instability, especially on unanswerable queries, making both its runtime and output quality unpredictable. AGRAG significantly reduces this instability and improves latency, but overall there is a clear trade-off: RAG is fastest but weakest, GRAG is unreliable, AGRAG balances cost and quality, and HRAG provides the most robust and consistent results.
\end{tcolorbox}

\subsection{Representative Failure Cases}
\label{subsec:Representative_Failure_Cases}

To illustrate the operational implications of our quantitative findings, we present 
three representative failure cases extracted from evaluation logs, demonstrating the 
real-world consequences of different retrieval architectures.

\paragraph{Case 1: Agentic Correction Recovers with Dramatic Speedup}




\begin{table}[t]
\centering
\small

\caption{Failure Case 1.}
\label{table:rep_fcase_1}
\begin{tabular}{p{0.18\linewidth} p{0.75\linewidth}}
\textbf{Question} & Which CVE was exploited in the APT37 Internet Explorer incident? \\
\textbf{Gold} & CVE-2022-41128 \\
\hline
\textbf{GRAG (736.4s)} & [Empty response after 12+ minutes] \\
\textit{Failure} & Generated a Cypher query with an incorrect relationship name.
Subsequent retry attempts continued searching for a schema pattern that does not exist, exhausting 25 correction iterations. \\
\hline
\textbf{AGRAG (4.9s)} & ``CVE-2022-41128'' \checkmark \\
\textit{Mechanism} & Critique loop identified relationship schema mismatch in the first iteration and regenerated a valid query. \\
\end{tabular}
\vspace{-3mm}
\end{table}

In this case, the system is asked to identify the CVE exploited in the APT37 Internet Explorer incident, which requires mapping the incident to its associated CVE in the graph schema. This 147× speedup (736s → 5s) (see Table~\ref{table:rep_fcase_1}) demonstrates that agentic correction not only improves 
accuracy but \emph{reduces} latency by detecting errors before entering prolonged retry 
loops. The critique mechanism prevented the pathological behavior observed in GRAG, 
where iterative validation attempts variations of an invalid query pattern until timeout.

\paragraph{Case 2: Hybrid Retrieval Enables Safe Abstention}





\begin{table}[t]
\centering
\small
\caption{Failure Case 2.}
\label{table_rep_fcase_2}
\begin{tabular}{p{0.18\linewidth} p{0.75\linewidth}}
\textbf{Question} & What CVSS score is recorded for CVE-2024-21338? \\
\textbf{Gold} & [Information not available in reports] \\
\hline
\textbf{RAG (1.2s)} & ``The CVSS score recorded for CVE-2024-21338 is 9.1'' \\
\textit{Safety failure} & Confidently hallucinated a specific numerical value that is not present in the source documents. \\
\hline
\textbf{GRAG (520.4s)} & [Empty response after 8.7-minute timeout] \\
\textit{Partial success} & Did not produce an answer, but only after prolonged latency while searching for a non-existent CVE property in the graph. \\
\hline
\textbf{HRAG (15.3s)} & ``I don't know'' \checkmark \\
\textit{Correct behavior} & Correctly identified missing information through cross-validation: the graph query returned no CVE node with a CVSS property and text retrieval found no CVSS mention. \\
\end{tabular}
\vspace{-3mm}
\end{table}

This case (see Table~\ref{table_rep_fcase_2}) exemplifies the safety-latency trade-off inherent in different retrieval 
architectures. RAG provides immediate but false information, prioritizing speed over 
correctness. GRAG eventually abstains but only after unacceptable latency. Only HRAG 
achieves both rapid response (15s) and correct refusal through redundant evidence paths.

\paragraph{Case 3: Query Correction Death Spiral}

The most extreme failure occurred on the unanswerable query ``What exact GoAnywhere MFT 
version is vulnerable to CVE-2024-0204?'' GRAG entered a pathological correction loop 
lasting 2,348 seconds (39.1 minutes), iteratively generating invalid Cypher 
queries searching for a property that does not exist in the schema, before timing out 
with an empty result.

Detailed log analysis reveals the failure mechanism:
\begin{enumerate}
\item Initial Cypher query searches for \texttt{CVE.affected\_version} property
\item Query returns empty result (property does not exist in schema)
\item Correction loop attempts alternative property names: \texttt{version}, 
\texttt{vulnerable\_version}, \texttt{product\_version}
\item All attempts fail validation or return empty results
\item Process repeats for 25 correction iterations over 39 minutes
\item System finally returns empty response after timeout
\end{enumerate}

For operational CTI settings, where analysts often require near real-time responses during incident triage, a response latency of 39 minutes for a non-answer ("information not available") is operationally impractical. This is not an implementation artifact but a fundamental limitation: without explicit uncertainty modeling, schema-constrained generation cannot represent ``unknown'' and instead exhausts retry budgets searching for impossible 
graph paths.

\textbf{Comparison to other systems:} On this same query, RAG hallucinated a specific version number in 1.3s (fast but wrong), AGRAG also produced an answer instead of abstaining (fast but unsafe), and HRAG correctly abstained in 18.7s (acceptable latency, correct refusal).
Only GRAG exhibited the timeout behavior.

\subsection{Adversary Model: Systematic Failure Analysis}
\label{sec:adversary}

To move beyond aggregate performance metrics, we construct a structured adversary model that systematically characterizes how a strategic adversary, or equivalently, a challenging real-world query distribution, can exploit failure modes 
in each RAG architecture. This analysis is relevant for CTI deployment, where adversaries may craft queries to induce incorrect analyst decisions, and where naturally occurring edge cases (ambiguous, underspecified, or out-of-scope questions) 
produce equivalent failure patterns.

\subsubsection{Threat Model}

We consider an adversary whose goal is to induce decision error in a CTI analyst who relies on a RAG-based assistant. The adversary can influence the natural-language query but cannot modify the knowledge base, the retrieval infrastructure, or the LLM. We define a successful attack as inducing a response with a judge score $\leq$15/50, indicating a substantially wrong, unfaithful, or unhelpful answer. The adversary may have varying levels of knowledge:

\begin{itemize}
    \item \textbf{Black-box:} The adversary knows only that a RAG system is deployed but not which architecture variant.
    \item \textbf{Grey-box:} The adversary knows which architecture variant is deployed (RAG, GRAG, AGRAG, or HRAG).
    \item \textbf{White-box:} The adversary additionally knows the graph schema and can craft queries that target schema gaps.
\end{itemize}

\subsubsection{Attack Strategies and Success Rates}

We evaluate four attack strategies derived from the question taxonomy, plus three compound strategies combining question properties. Table~\ref{tab:attack_surface} reports the conditional probability of attack success (judge score~$\leq$15) 
under each strategy.

\begin{table*}[t]
\centering
\caption{Adversary attack success rates (\% of questions scoring $\leq$15/50) 
by strategy and system. Higher values indicate greater vulnerability. Bold marks 
the most effective attack per system.}
\label{tab:attack_surface}
\begin{tabular}{p{6.0cm}rrrr}
\toprule
\textbf{Attack Strategy} & \textbf{RAG} & \textbf{GRAG} & \textbf{AGRAG} & \textbf{HRAG} \\
\midrule
\multicolumn{5}{l}{\textit{Single-category strategies}} \\
A1: Schema evasion (guided)    
& 20.6\% & \textbf{82.5\%} & 18.1\% &  3.1\% \\
A2: Unanswerable probing       
&  6.0\% & 46.0\% &  4.0\% & \textbf{20.0\%} \\
A3: Multi-hop stress            
& 47.3\% & 16.7\% & 18.7\% & 12.0\% \\
A4: Simple fact (control)       
& \textbf{52.0\%} & 11.3\% & 12.0\% &  6.7\% \\
\midrule
\multicolumn{5}{l}{\textit{Compound strategies}} \\
A5: Multi-entity + guided       
& 20.8\% & 83.0\% & 18.2\% &  3.1\% \\
A6: Long questions ($>75$th pctl) 
& 21.5\% & 81.6\% & 17.8\% &  3.1\% \\
A7: Aggregate multi-hop         
& 44.4\% & 15.3\% & \textbf{26.4\%} & \textbf{22.2\%} \\
\midrule
Random (any)           
& 37.0\% & 33.3\% & 13.2\% &  6.7\% \\
\bottomrule
\end{tabular}
\vspace{-4mm}
\end{table*}

\paragraph{Key findings.}
Each system exhibits a distinct weakness that a grey-box adversary can exploit:

\begin{itemize}
    \item \textbf{RAG} is most vulnerable to simple fact lookups (A4: 52.0\% attack success), where semantic retrieval fails to surface the specific entity or property. This counterintuitive result, simple questions being harder than complex ones, arises because RAG's top-$k$ chunk retrieval frequently misses short, precise facts embedded in longer narrative passages.
    
    \item \textbf{GRAG} is catastrophically vulnerable to schema evasion (A1: 82.5\%), 
    where guided analyst-style questions reference concepts outside the graph schema. 
    The attack success rate increases further with multi-entity questions (A5: 83.0\%) 
    and long questions (A6: 81.6\%), as these amplify the probability of referencing 
    at least one out-of-schema concept.
    
    \item \textbf{AGRAG} is most vulnerable to aggregate multi-hop queries 
    (A7: 26.4\%), which require compositional Cypher (e.g., \texttt{COUNT}, 
    \texttt{COLLECT}) that agentic critique cannot always repair.
    
    \item \textbf{HRAG} achieves the lowest vulnerability across most strategies, with 
    its main weakness being unanswerable probing (A2: 20.0\%) and aggregate multi-hop 
    (A7: 22.2\%).
\end{itemize}

A grey-box adversary gains substantial advantage over random querying. 
For GRAG, a targeted attack using guided questions raises the attack success rate from 
33.3\% (random) to 82.5\% (schema evasion), a 2.5$\times$ amplification. For RAG, 
targeting simple facts increases success from 37.0\% to 52.0\% (1.4$\times$). 
HRAG is the most resistant to strategic targeting: even the best attack (A2/A7) 
achieves only 20-22\%, compared to 6.7\% under random queries (3.0$\times$ amplification), 
making it the hardest system to exploit.

\subsubsection{Failure Mode Classification}

\begin{table}[t]
\centering
\caption{Failure mode classification across RAG systems. 
Modes are defined by judge criterion scores: Hallucination = C3$\leq$2 and C4$\geq$3 
(fluent but unfaithful); Full Collapse = C1, C2, C3 all $\leq$2.}
\label{tab:failure_modes}
\begin{tabular}{lrrrr}
\toprule
\textbf{Failure Mode} & \textbf{RAG} & \textbf{GRAG} & \textbf{AGRAG} & \textbf{HRAG} \\
\midrule
Correct ($\geq$40)    & 41.5\% & 61.1\% & 74.2\% & 83.2\% \\
Partial (10--40)       & 31.7\% & 12.3\% & 14.1\% & 12.0\% \\
Near-zero ($\leq$10)   & 26.8\% & 26.7\% & 11.7\% &  4.8\% \\
\midrule
Hallucination          & 45.6\% & 28.2\% &  9.7\% & 11.7\% \\
Full collapse          & 41.1\% & 33.0\% & 15.8\% &  8.8\% \\
\bottomrule
\end{tabular}
\vspace{-5mm}
\end{table}

Beyond binary success/failure, Table~\ref{tab:failure_modes} classifies outputs into 
distinct failure modes. RAG fails primarily through hallucination (45.6\% of all 
answers are fluent but unfaithful), while GRAG fails through full collapse (33.0\% 
with all correctness criteria $\leq$2), typically caused by empty Cypher results or 
timeout. AGRAG and HRAG reduce both failure modes substantially, with HRAG achieving 
the lowest rates across all categories.

Critically, the two dominant failure modes, hallucination and collapse, require different 
mitigations. Hallucination (RAG's primary mode) requires improved retrieval precision or 
faithfulness constraints. Collapse (GRAG's primary mode) requires bounded execution and 
fallback paths. AGRAG addresses collapse through critique-based early exit; HRAG addresses 
both through redundant evidence paths.

\subsubsection{Timing Side Channel}

\begin{table}[t]
\centering

\begin{minipage}[t]{0.43\textwidth}
\centering
\caption{GRAG timing as a failure signal. Response time $>T$ seconds predicts 
a failed answer (score $\leq$10). Precision = \% of slow responses that are failures; 
Recall = \% of failures that are slow.}
\label{tab:timing_side_channel}
\begin{tabular}{cccc}
\toprule
\textbf{Threshold $T$ (s)} & \textbf{$n$ (above)} & \textbf{Precision} & \textbf{Recall} \\
\midrule
30  & 292 & 50\% & 84\% \\
60  & 188 & 56\% & 60\% \\
120 & 135 & 61\% & 47\% \\
300 &  73 & 70\% & 29\% \\
500 &  48 & 77\% & 21\% \\
\bottomrule
\end{tabular}
\end{minipage}
\hfill
\begin{minipage}[t]{0.50\textwidth}
\centering
\caption{Ensemble defense: failure rates under oracle selection (best score 
among the subset of systems). Fail = judge score $\leq$10.}
\label{tab:ensemble_defense}
\small
\begin{tabular}{p{3.0cm}ccc}
\toprule
\textbf{System configuration} & \textbf{Mean score} & \textbf{Fail} & \textbf{Perfect} \\
\midrule
RAG alone                  & 29.4 & 26.8\% & 31.7\% \\
GRAG alone                 & 34.1 & 26.7\% & 55.0\% \\
AGRAG alone                & 40.7 & 11.7\% & 69.5\% \\
HRAG alone                 & 44.7 &  4.8\% & 74.8\% \\
\midrule
Best(AGRAG, HRAG)          & 47.9 &  0.9\% & 86.5\% \\
Best(RAG, HRAG)            & 46.6 &  1.5\% & 79.4\% \\
Best(all four)             & 49.0 &  0.0\% & 90.2\% \\
\bottomrule
\end{tabular}
\end{minipage}

\vspace{-4mm}
\end{table}


GRAG's iterative query-correction loops create a measurable timing side channel: 
failed queries take significantly longer than successful ones (mean 292.6s vs.\ 45.0s, 
a 6.5$\times$ ratio). Table~\ref{tab:timing_side_channel} shows that GRAG response 
time can serve as a failure detector: at a 60-second threshold, 56\% of slow responses 
are failures (precision) and 60\% of all failures are slow (recall). At 300 seconds, 
precision reaches 70\%. This timing correlation has two security implications. First, an adversary observing 
response latency can infer whether the system failed to answer, even without seeing 
the response content. Second, the timing signal can be exploited defensively: a 
deployment could impose a hard timeout and return ``insufficient information'' rather 
than waiting for a potentially low-quality answer after prolonged retry loops. Notably, the other systems do not exhibit this vulnerability. RAG shows no timing-failure correlation ($r{=}{-}0.09$), HRAG shows negligible correlation ($r{=}{-}0.02$), and 
AGRAG's correlation is moderate ($r{=}{+}0.30$) but with bounded absolute latency (failure mean: 17.9s vs.\ success: 8.3s).

\subsubsection{Failure Decorrelation and Ensemble Defense}


A key question for defensive deployment is whether system failures are correlated, if 
all systems fail on the same questions, redundancy provides no benefit. 
Table~\ref{tab:failure_correlation} shows that failures are largely decorrelated 
across architectures. The Pearson correlation between RAG and GRAG failure indicators 
is negative ($r{=}{-}0.22$), meaning they tend to fail on different 
questions. AGRAG-HRAG failure correlation is low ($r{=}0.08$), and all cross-architecture 
Jaccard overlaps are below 0.15 (Table~\ref{tab:failure_jaccard}).


\begin{table*}[t]
\centering

\begin{minipage}[t]{0.5\linewidth}
\centering
\caption{Failure decorrelation: Pearson correlation of binary failure indicators (score $\leq$15 = fail).}
\label{tab:failure_correlation}
\begin{tabular}{lcccc}
\toprule
 & \textbf{RAG} & \textbf{GRAG} & \textbf{AGRAG} & \textbf{HRAG} \\
\midrule
RAG   & 1.000 & $-$0.215 & 0.008 & 0.034 \\
GRAG  &       & 1.000    & 0.105 & $-$0.034 \\
AGRAG &       &          & 1.000 & 0.075 \\
HRAG  &       &          &       & 1.000 \\
\bottomrule
\end{tabular}
\end{minipage}
\hfill
\begin{minipage}[t]{0.45\linewidth}
\centering
\caption{Jaccard overlap of failure sets (shared / union).}
\label{tab:failure_jaccard}
\begin{tabular}{lc}
\toprule
\textbf{Pair} & \textbf{Jaccard} \\
\midrule
RAG--GRAG   & 0.118 \\
AGRAG--HRAG & 0.083 \\
RAG--HRAG   & 0.071 \\
GRAG--HRAG  & 0.048 \\
\bottomrule
\end{tabular}
\end{minipage}

\end{table*}

This decorrelation enables effective ensemble defense. Table~\ref{tab:ensemble_defense} shows that oracle selection between AGRAG and HRAG reduces the failure rate from 4.8\% (HRAG alone) to 0.9\%, while raising the 
perfect-score rate to 86.5\%. The oracle over all four systems achieves zero failures (0.0\%) and 90.2\% perfect scores. While oracle selection requires a reliable meta-classifier, these results establish an upper bound for 
ensemble-based defenses and demonstrate that the residual failure surfaces of different architectures are largely complementary.

\subsubsection{Question Features that Predict Failure}

\begin{table}[t]
\centering
\caption{Correlation between question-level features and failure 
(score $\leq$15). Positive values indicate the feature increases failure risk.}
\label{tab:question_features}
\begin{tabular}{p{5.0cm}cccc}
\toprule
\textbf{Feature} & \textbf{RAG} & \textbf{GRAG} & \textbf{AGRAG} & \textbf{HRAG} \\
\midrule
Question length (words) & $-$0.19 & $+$0.58 & $+$0.09 & $-$0.07 \\
Multi-entity (and/or/all) & $-$0.17 & $+$0.52 & $+$0.05 & $-$0.09 \\
Gold answer length (words)& $-$0.16 & $+$0.54 & $+$0.05 & $-$0.07 \\
Contains ``which''         & $-$0.15 & $+$0.22 & $-$0.04 & $-$0.17 \\
Contains ``how many''      & $+$0.01 & $-$0.05 & $+$0.13 & $+$0.17 \\
Has temporal reference     & $+$0.08 & $-$0.03 & $+$0.00 & $-$0.01 \\
\bottomrule
\end{tabular}
\vspace{-5mm}
\end{table}

To characterize which question properties a white-box adversary could exploit, 
Table~\ref{tab:question_features} reports correlations between surface-level question 
features and failure. Two patterns emerge. First, RAG and GRAG exhibit 
inverted vulnerability profiles: question length, multi-entity references, and 
complex gold answers \emph{reduce} RAG failure risk (negative correlation) but 
\emph{strongly increase} GRAG failure risk ($r{=}+0.52$ to $+0.58$). This occurs 
because longer, more complex questions tend to be guided analyst-style questions, 
which RAG handles through broad retrieval but GRAG cannot express within its schema.

Second, aggregate queries (``how many'', ``count'') are the strongest 
predictors of AGRAG and HRAG failure ($r{=}+0.13$ and $+0.17$), reflecting the 
difficulty of generating correct aggregate Cypher even with agentic correction.

These inverted profiles confirm that a white-box adversary must tailor attacks to 
the deployed architecture: long, multi-entity questions attack GRAG; short, 
specific fact queries attack RAG; aggregate counting queries stress AGRAG and HRAG.

\subsubsection{Adversary Model Summary}

\begin{table}[t]
\centering
\caption{Adversary model summary: primary vulnerability, dominant failure mode, 
best-case attack success rate, and recommended mitigation per system.}
\label{tab:adversary_summary}
\begin{tabular}{p{1cm}p{3cm}p{3cm}p{2cm}p{4cm}}
\toprule
\textbf{System} & \textbf{Primary Vulnerability} & \textbf{Failure Mode} & \textbf{Best Attack} & \textbf{Mitigation} \\
\midrule
RAG & Simple fact lookup & Hallucination (45.6\%) & 52.0\% & Improve retrieval precision \\
GRAG & Schema evasion & Full collapse (33.0\%) & 82.5\% & Bounded timeout + fallback \\
AGRAG & Aggregate queries & Partial failure (14.1\%) & 26.4\% & Aggregate query templates \\
HRAG & Unanswerable probing & Low (8.8\% collapse) & 22.2\% & Explicit abstention logic \\
\bottomrule
\end{tabular}
\vspace{-5mm}
\end{table}

Table~\ref{tab:adversary_summary} consolidates the adversary analysis. The four 
architectures exhibit fundamentally different failure profiles, which has 
two implications for CTI deployment:

\begin{enumerate}
    \item \textbf{No single system is robust to all attack strategies.} RAG is 
    vulnerable to fact-seeking queries, GRAG to out-of-schema queries, AGRAG to 
    aggregate reasoning, and HRAG to unanswerable probing. Deployment decisions 
    should match the expected query distribution to the system's strength profile.
    
    \item \textbf{Failure decorrelation enables ensemble defense.} Because 
    architectures fail on different questions (Jaccard overlap $<$0.15), combining 
    even two systems (e.g., AGRAG + HRAG) reduces the failure rate from 4.8\% to 
    0.9\%. A meta-classifier that routes queries to the most appropriate system, 
    or a voting mechanism that detects disagreement, could approach the oracle 
    upper bound of 0.0\% failure.
\end{enumerate}

These findings extend this work's security-relevant failure analysis from 
descriptive observation to a predictive adversary model, providing actionable 
guidance for hardening CTI assistants against both strategic attacks and 
naturally occurring edge cases.

\section{Discussion}
\label{sec:discussion}
In this section we present this paper's overview, analyses and the findings. 
\subsection{Overview}

This study investigates whether graph-based retrieval architectures
improve question answering over CTI reports. Across 3,300 questions and five language models, the results reveal a consistent pattern: while explicit graph grounding can significantly improve performance for structured factual queries, naive graph-only pipelines introduce new failure modes that can outweigh their advantages. In practice, the benefits of graph retrieval only emerge when combined with agentic query correction or hybrid retrieval redundancy. 

\subsection{Analyses}
Graph grounding provides clear advantages for fact-centric CTI questions, particularly those requiring relational reasoning. For simple, single-hop, and multi-hop questions, AGRAG and HRAG consistently outperform vanilla semantic retrieval
(Table~\ref{tab:rq2_judge_by_category}). These improvements are most pronounced for multi-hop reasoning, where answers depend on chaining relationships across entities. In such cases, semantic retrieval often fails to retrieve all relevant passages simultaneously, whereas graph traversal can directly follow relational paths. However, graph grounding alone is not sufficient. Plain GraphRAG yields only marginal average improvements despite producing more perfect answers. This indicates that the reliability of the text-to-query translation layer is a critical bottleneck for graph-based retrieval. Graph-only retrieval introduces failure modes that are largely absent in semantic RAG systems. Because answering requires translating natural language into executable Cypher queries, errors in query generation, schema matching, or result interpretation can cause the system to fail. These failures are particularly evident for questions outside the graph schema. Guided analyst-style queries often reference contextual information not captured in the structured graph, leading GRAG to produce incorrect answers or enter prolonged query-repair loops. Similarly, unanswerable questions expose a structural limitation: when no supporting graph path exists, graph-only systems tend to repeatedly generate alternative queries rather than abstaining. We refer to this phenomenon as \emph{structural hallucination}, where the system produces answers consistent with the graph query output but not supported by the underlying reports due to incomplete schema coverage. 

\subsection{Findings}
Our results demonstrate that architectural safeguards are essential for reliable graph-based retrieval. Agentic correction substantially reduces query-generation failures by detecting and repairing invalid Cypher queries early. Hybrid retrieval provides an even stronger mitigation by combining graph traversal with semantic evidence retrieval, allowing the system to fall back on unstructured context when structured queries fail. From a system design perspective, these findings suggest that graph retrieval should not replace semantic retrieval but complement it within a broader retrieval architecture. For operational CTI workflows, the primary risk of AI assistants is not average performance but catastrophic failures on edge cases. Graph-only retrieval is particularly vulnerable to such failures, as schema gaps and query-generation errors can produce unstable or delayed responses. Hybrid architectures offer a more robust solution by combining complementary evidence sources and enabling safer abstention when information is missing.

\begin{table}[t]
\centering
\caption{SWOT analysis of evaluated RAG architectures for CTI question answering.}
\label{tab:swot_rag}
\small
\begin{tabular}{p{1.0cm}p{3.1cm}p{3.1cm}p{3.1cm}p{3.1cm}}
\toprule
\textbf{System} & \textbf{Strengths} & \textbf{Weaknesses} & \textbf{Opportunities} & \textbf{Threats} \\
\midrule

\textbf{RAG} &
Fast and low-cost retrieval; robust to schema gaps; stable runtime behavior. &
High hallucination rates; weak multi-hop reasoning; sensitive to top-$k$ retrieval failures. &
Improved reranking or hybrid retrieval could enhance factual recall. &
Vulnerable to simple fact queries and overconfident answers for missing information. \\

\textbf{GRAG} &
Explicit relational reasoning over CTI entities; strong factual lookup when Cypher queries succeed. &
Highly sensitive to text-to-Cypher errors; latency instability due to retry loops. &
Improved text-to-query models and bounded correction strategies could stabilize performance. &
Schema evasion attacks and catastrophic collapse on unanswerable queries. \\

\textbf{AGRAG} &
Agentic critique improves Cypher reliability; large accuracy gains over classical RAG; reduces query failures. &
Higher latency than RAG; dependent on graph schema coverage. &
Advanced agentic planning and better query validation could further improve robustness. &
Aggregate queries and schema gaps remain challenging. \\

\textbf{HRAG} &
Best overall robustness; combines graph reasoning with semantic retrieval; lowest hallucination rates. &
Higher computational cost and system complexity. &
Adaptive routing or ensemble strategies could reduce cost while preserving robustness. &
Incorrect pipeline balancing or infrastructure constraints may limit deployment. \\

\bottomrule
\end{tabular}
\vspace{-6mm}
\end{table}

To synthesize the architectural trade-offs observed in our evaluation, Table~\ref{tab:swot_rag} summarizes the strengths, weaknesses, opportunities, and threats (SWOT) of the four evaluated retrieval architectures.

\subsection{Recommendations}

These findings suggest several practical recommendations. For system deployment, hybrid retrieval pipelines should be preferred over
graph-only architectures, and systems should include safeguards such as bounded query-repair loops and explicit abstention mechanisms.
For research, the main challenge lies in improving the reliability of text-to-query generation and developing evaluation benchmarks that
capture realistic failure modes such as schema gaps and unanswerable queries.

\subsection{Limitations and Future Work}

This study focuses on a single CTI dataset and a controlled graph schema derived from CTINexus reports. Different datasets or denser relational graphs may lead to different trade-offs between structured and semantic retrieval. Additionally, the evaluation uses automatically generated questions, which may not fully capture the diversity of analyst queries. Future work should therefore include human analyst studies, improved uncertainty modeling for safe abstention, and robustness evaluations under incomplete or adversarial CTI data. Another promising direction is the development of adaptive routing or ensemble architectures that dynamically select or combine retrieval pipelines based on query characteristics, potentially reducing computational cost while preserving the robustness benefits of hybrid retrieval.

\section{Conclusion}
\label{sec:conclusion}
We presented a controlled evaluation of four RAG architectures for cyber threat intelligence: vanilla semantic RAG, graph-only GraphRAG, agentic GraphRAG, and a hybrid graph+text system (HRAG). Across 3{,}300 CTI QA pairs and five LLMs, we find that explicit graph grounding alone is not a reliable upgrade: plain GraphRAG yields only marginal quality gains and exhibits severe latency instability due to query-repair loops. In contrast, agentic refinement and hybrid redundancy deliver large, consistent improvements, especially for structured fact-seeking and multi-hop queries, while also reducing catastrophic text-to-Cypher failures. We further show that the effectiveness of graph-based retrieval is strongly model-dependent and that safe behavior on unanswerable queries remains a key challenge for graph-centric systems.

Our evaluation logs reveal that the primary risk of graph-based CTI retrieval is not average-case performance degradation, but catastrophic failure modes on edge cases: 100\% collapse rate on unanswerable queries, multi-minute timeout loops (up to 39 minutes observed), and overconfident answers from partial graph structure. These findings demonstrate that graph grounding alone is insufficient - reliable CTI assistants require explicit uncertainty modeling, bounded correction mechanisms (AGRAG achieves up to 147× speedup over GRAG), or hybrid redundancy (HRAG achieves 76\% correct refusal vs. 0\% for the semantic RAG) to ensure safe operation under realistic query distributions.

Overall, our results suggest that graph-based retrieval should be treated as a high-risk component in CTI assistants: it requires agentic correction mechanisms and/or parallel unstructured text retrieval, like in the HRAG, to realize its benefits without sacrificing robustness. Future work should incorporate analyst-in-the-loop evaluation, explicit abstention and evidence-coverage checks, and robustness testing under realistic data quality issues and adversarial conditions.

\begin{acks}
Funded by the European Union under the Horizon Europe Research and Innovation programme (GA no. 101168144 - MIRANDA). Views and opinions expressed are however those of the author(s) only and do not necessarily reflect those of the European Union. Neither the European Union nor the granting authority can be held responsible for them.
\end{acks}

\bibliographystyle{ACM-Reference-Format}
\bibliography{sample-base-new}

@String{Computing = "Computing" }

@String{Springer = "Springer-Verlag" }

@INPROCEEDINGS{Cheng2025_official,
  author={Cheng, Yutong and et al.},
  booktitle={2025 IEEE 10th European Symposium on Security and Privacy (EuroS\&P)}, 
  title={CTINexus: Automatic Cyber Threat Intelligence Knowledge Graph Construction Using Large Language Models}, 
  year={2025},
  volume={},
  number={},
  pages={923-938},
  doi={10.1109/EuroSP63326.2025.00057}}

@inproceedings{huang-etal-2025-selfaug,
    title = "{S}elf{A}ug: Mitigating Catastrophic Forgetting in Retrieval-Augmented Generation via Distribution Self-Alignment",
    author = {Huang, Yuqing and et al.},
    booktitle = "Findings of the Association for Computational Linguistics: EMNLP 2025",
    month = nov,
    year = "2025",
    address = "Suzhou, China",
    publisher = "Association for Computational Linguistics",
    url = "https://aclanthology.org/2025.findings-emnlp.763/",
    doi = "10.18653/v1/2025.findings-emnlp.763",
    pages = "14175--14190",
    ISBN = "979-8-89176-335-7"
}

@article{zheng2023judging,
  title={Judging LLM-as-a-Judge with MT-Bench and Chatbot Arena},
  author={Zheng, Lianmin and et al.},
  journal={arXiv preprint arXiv:2306.05685},
  year={2023}
}

@inproceedings{blagec-etal-2022-global,
    title = "A global analysis of metrics used for measuring performance in natural language processing",
    author = {Blagec, Kathrin and et al.},
    booktitle = "Proceedings of NLP Power! The First Workshop on Efficient Benchmarking in NLP",
    month = may,
    year = "2022",
    address = "Dublin, Ireland",
    publisher = "Association for Computational Linguistics",
    url = "https://aclanthology.org/2022.nlppower-1.6/",
    doi = "10.18653/v1/2022.nlppower-1.6",
    pages = "52--63"
}

@inproceedings{lin-2004-rouge,
    title = "{ROUGE}: A Package for Automatic Evaluation of Summaries",
    author = "Lin, Chin-Yew",
    booktitle = "Text Summarization Branches Out",
    month = jul,
    year = "2004",
    address = "Barcelona, Spain",
    publisher = "Association for Computational Linguistics",
    url = "https://aclanthology.org/W04-1013/",
    pages = "74--81"
}

@misc{mehra2025improvingapplicabilitydeeplearning,
      title={Improving Applicability of Deep Learning based Token Classification models during Training}, 
      author={Mehra, Anket and et al.},
      year={2025},
      eprint={2504.01028},
      archivePrefix={arXiv},
      primaryClass={cs.CV},
      url={https://arxiv.org/abs/2504.01028}, 
}

@misc{zhang2020bertscoreevaluatingtextgeneration,
      title={BERTScore: Evaluating Text Generation with BERT}, 
      author={Zhang, Tianyi and et al.},
      year={2020},
      eprint={1904.09675},
      archivePrefix={arXiv},
      primaryClass={cs.CL},
      url={https://arxiv.org/abs/1904.09675}, 
}

@article{liu2023geval,
  title={G-Eval: NLG Evaluation using GPT-4 with Better Human Alignment},
  author={Liu, Yang and et al.},
  journal={arXiv preprint arXiv:2303.16634},
  year={2023}
}

@article{fu2023gptscore,
  title={GPTScore: Evaluate as You Desire},
  author={Fu, Jinlan and et al.},
  journal={arXiv preprint arXiv:2302.04166},
  year={2023}
}

@misc{barres2025tau2benchevaluatingconversationalagents,
      title={$\tau^2$-Bench: Evaluating Conversational Agents in a Dual-Control Environment}, 
      author={Barres, Victor and et al.},
      year={2025},
      eprint={2506.07982},
      archivePrefix={arXiv},
      primaryClass={cs.AI},
      url={https://arxiv.org/abs/2506.07982}, 
}

@preprint{Gu2025,
       author = {Gu, Jiawei and et al.},
        title = {A Survey on LLM-as-a-Judge},
         year = 2025,
archivePrefix = {arXiv},
       eprint = {2411.15594},
 primaryClass = {cs.CL},
          url = {https://arxiv.org/abs/2411.15594},
}

@preprint{FAIRCodeGen2025,
       author = {{FAIR CodeGen team} and et al.},
        title = {CWM: An Open-Weights LLM for Research on Code Generation with World Models},
         year = 2025,
          eid = {arXiv:2510.02387},
archivePrefix = {arXiv},
       eprint = {2510.02387},
 primaryClass = {cs.SE},
}

@article{angles2018propertygraph,
  title={Foundations of Modern Graph Query Languages},
  author={Angles, Renzo and et al.},
  journal={ACM Computing Surveys},
  volume={50},
  number={5},
  pages={1--40},
  year={2018},
  publisher={ACM},
  doi={10.1145/3104031}
}

@inproceedings{sun-etal-2019-pullnet,
    title = "{P}ull{N}et: Open Domain Question Answering with Iterative Retrieval on Knowledge Bases and Text",
    author = {Sun, Haitian and et al.},
    booktitle = "Proceedings of the 2019 Conference on Empirical Methods in Natural Language Processing and the 9th International Joint Conference on Natural Language Processing (EMNLP-IJCNLP)",
    month = nov,
    year = "2019",
    address = "Hong Kong, China",
    publisher = "Association for Computational Linguistics",
    url = "https://aclanthology.org/D19-1242/",
    doi = "10.18653/v1/D19-1242",
    pages = "2380--2390"
}

@inproceedings{nogueira2019passage,
  title={Passage Re-ranking with BERT},
  author={Nogueira, Rodrigo and Cho, Kyunghyun},
  booktitle={Proceedings of the 2019 Conference on Empirical Methods in Natural Language Processing (EMNLP)},
  year={2019}
}

@misc{jiang2024longragenhancingretrievalaugmentedgeneration,
      title={LongRAG: Enhancing Retrieval-Augmented Generation with Long-context LLMs}, 
      author={Jiang, Ziyan and et al.},
      year={2024},
      eprint={2406.15319},
      archivePrefix={arXiv},
      primaryClass={cs.CL},
      url={https://arxiv.org/abs/2406.15319}, 
}

@inproceedings{wang-etal-2025-document,
    title = "Document Segmentation Matters for Retrieval-Augmented Generation",
    author = {Wang, Zhitong and et al.},
    booktitle = "Findings of the Association for Computational Linguistics: ACL 2025",
    month = jul,
    year = "2025",
    address = "Vienna, Austria",
    publisher = "Association for Computational Linguistics",
    url = "https://aclanthology.org/2025.findings-acl.422/",
    doi = "10.18653/v1/2025.findings-acl.422",
    pages = "8063--8075",
    ISBN = "979-8-89176-256-5",
}

@inproceedings{asai2019learning,
  title={Learning to Retrieve Reasoning Paths over Wikipedia Graph for Question Answering},
  author={Asai, Akari and et al.},
  booktitle={International Conference on Learning Representations (ICLR)},
  year={2020}
}

@article{gao2023retrieval,
  title={Retrieval-Augmented Generation for Large Language Models: A Survey},
  author={Gao, Yunfan and et al.},
  journal={arXiv preprint arXiv:2312.10997},
  year={2023}
}

@misc{edge2025graphrag,
  title        = {From Local to Global: A {GraphRAG} Approach to Query-Focused Summarization},
  author       = {Edge, Darren and Trinh, Ha and Cheng, Newman and Bradley, Joshua and Chao, Alex and Mody, Apurva and Truitt, Steven and Metropolitansky, Dasha and Ness, Robert Osazuwa and Larson, Jonathan},
  year         = {2025},
  eprint       = {2404.16130},
  archivePrefix= {arXiv},
  primaryClass = {cs.CL},
  note         = {arXiv:2404.16130v2}
}

@misc{pan2024unifying,
  title        = {Unifying Large Language Models and Knowledge Graphs: A Roadmap},
  author       = {Pan, Shirui and Luo, Linhao and Wang, Yufei and Chen, Chen and Wang, Jiapu and Wu, Xindong},
  year         = {2024},
  eprint       = {2306.08302},
  archivePrefix= {arXiv},
  primaryClass = {cs.CL},
  note         = {arXiv:2306.08302v3}
}

@misc{liang2025graphragunderfire,
  title        = {{GraphRAG} under Fire},
  author       = {Liang, Jiacheng and Wang, Yuhui and Li, Changjiang and Zhu, Rongyi and Jiang, Tanqiu and Gong, Neil and Wang, Ting},
  year         = {2025},
  eprint       = {2501.14050},
  archivePrefix= {arXiv},
  primaryClass = {cs.LG},
  note         = {arXiv:2501.14050v1}
}

@misc{xiang2025whentouse,
  title        = {When to use Graphs in {RAG}: A Comprehensive Analysis for Graph Retrieval-Augmented Generation},
  author       = {Xiang, Zhishang and Wu, Chuanjie and Zhang, Qinggang and Chen, Shengyuan and Hong, Zijin and Huang, Xiao and Su, Jinsong},
  year         = {2025},
  eprint       = {2506.05690},
  archivePrefix= {arXiv},
  primaryClass = {cs.CL},
  note         = {arXiv:2506.05690v2}
}

@inproceedings{wu2025medicalgraphrag,
  title     = {Medical Graph {RAG}: Evidence-based Medical Large Language Model via Graph Retrieval-Augmented Generation},
  author = {Wu, Junde and et al.},
  booktitle = {Proceedings of the 63rd Annual Meeting of the Association for Computational Linguistics (Long Papers)},
  year      = {2025},
  address   = {Vienna, Austria},
  publisher = {Association for Computational Linguistics},
  pages     = {28443--28467},
  doi       = {10.18653/v1/2025.acl-long.1381},
  url       = {https://aclanthology.org/2025.acl-long.1381/}
}

@article{agentic_clinical_rag2025,
  title   = {A Self-Correcting Agentic Graph RAG Framework for Clinical Decision Support},
  author  = {Anonymous},
  journal = {Frontiers in Medicine},
  year    = {2025},
  note    = {Agentic RAG for clinical QA built on hepatology knowledge graph; significantly outperforms baseline RAG and GraphRAG} ,
  url     = {https://www.frontiersin.org/journals/medicine/articles/10.3389/fmed.2025.1716327/full}
}

@article{papageorgiou2025multimodal,
  title   = {Hybrid Multi-Agent GraphRAG for E-Government: Towards a Trustworthy AI Assistant},
  author  = {Papageorgiou, George and Sarlis, Vangelis and Maragoudakis, Manolis and Tjortjis, Christos},
  journal = {Applied Sciences},
  year    = {2025},
  volume  = {15},
  number  = {11},
  pages   = {6315},
  doi     = {10.3390/app15116315},
  url     = {https://www.mdpi.com/2076-3417/15/11/6315}
}

@inproceedings{lewis2020rag,
  title        = {Retrieval-Augmented Generation for Knowledge-Intensive {NLP} Tasks},
  author       = {Lewis, Patrick and Perez, Ethan and Piktus, Aleksandra and Petroni, Fabio and Karpukhin, Vladimir and Goyal, Naman and K{\"u}ttler, Heinrich and Lewis, Mike and Yih, Wen-tau and Rockt{\"a}schel, Tim and Riedel, Sebastian and Kiela, Douwe},
  booktitle    = {Advances in Neural Information Processing Systems},
  volume       = {33},
  year         = {2020},
  eprint       = {2005.11401},
  archivePrefix= {arXiv},
  primaryClass = {cs.CL},
  doi          = {10.48550/arXiv.2005.11401},
  url          = {https://arxiv.org/abs/2005.11401}
}

@misc{gao2024rag_survey,
  title        = {Retrieval-Augmented Generation for Large Language Models: A Survey},
  author       = {Gao, Yunfan and Xiong, Yun and Gao, Xinyu and Jia, Kangxiang and Pan, Jinliu and Bi, Yuxi and Dai, Yi and Sun, Jiawei and Wang, Meng and Wang, Haofen},
  year         = {2024},
  eprint       = {2312.10997},
  archivePrefix= {arXiv},
  primaryClass = {cs.CL},
  doi          = {10.48550/arXiv.2312.10997},
  url          = {https://arxiv.org/abs/2312.10997}
}

@article{ji2022hallucination,
  title        = {Survey of Hallucination in Natural Language Generation},
  author={Ji, Ziwei and et al.},
  journal      = {ACM Computing Surveys},
  year         = {2022},
  doi          = {10.1145/3571730},
  eprint       = {2202.03629},
  archivePrefix= {arXiv},
  primaryClass = {cs.CL},
  url          = {https://arxiv.org/abs/2202.03629}
}

@inproceedings{karpukhin2020dpr,
  title     = {Dense Passage Retrieval for Open-Domain Question Answering},
  author={Karpukhin, Vladimir and et al.},
  booktitle = {Proceedings of the 2020 Conference on Empirical Methods in Natural Language Processing (EMNLP)},
  year      = {2020},
  publisher = {Association for Computational Linguistics},
  doi       = {10.18653/v1/2020.emnlp-main.550},
  url       = {https://aclanthology.org/2020.emnlp-main.550/}
}

@inproceedings{rajpurkar2016squad,
  title     = {{SQuAD}: 100,000+ Questions for Machine Comprehension of Text},
  author    = {Rajpurkar, Pranav and Zhang, Jian and Lopyrev, Konstantin and Liang, Percy},
  booktitle = {Proceedings of the 2016 Conference on Empirical Methods in Natural Language Processing},
  pages     = {2383--2392},
  year      = {2016},
  publisher = {Association for Computational Linguistics}
}

@article{kwiatkowski2019natural,
  title     = {Natural Questions: A Benchmark for Question Answering Research},
  author={Kwiatkowski, Tom and et al.},
  journal   = {Transactions of the Association for Computational Linguistics},
  volume    = {7},
  pages     = {453--466},
  year      = {2019}
}

@article{peng2025graphrag,
  author={Peng, Boci and et al.},
  title = {Graph Retrieval-Augmented Generation: A Survey},
  journal = {ACM Transactions on Information Systems},
  volume = {44},
  number = {2},
  year = {2025},
  pages = {1--52},
  doi = {10.1145/3777378},
  publisher = {ACM},
  url = {https://doi.org/10.1145/3777378}
}

@article{OpenCyKG2021,
  title={Open-CyKG: An Open Cyber Threat Intelligence Knowledge Graph},
  author={Sarhan, Injy and Spruit, Marco},
  journal={Knowledge-Based Systems},
  year={2021},
  doi={10.1016/j.knosys.2021.107524}
}

@article{KnowCTI2024,
  title={KnowCTI: Knowledge-based cyber threat intelligence entity and relation extraction},
  author={Authors},
  journal={Journal of Computers \& Security},
  year={2024},
  note={Example CTI ontology with entities and relations},
  url={https://www.sciencedirect.com/science/article/pii/S0167404824001251}
}

@inproceedings{zhu2025kg2rag,
  author={Zhu, Xiangrong and et al.},
  title = {Knowledge Graph-Guided Retrieval Augmented Generation},
  booktitle = {Proceedings of the 2025 Conference of the Nations of the Americas Chapter of the Association for Computational Linguistics: Human Language Technologies (NAACL 2025)},
  series = {NAACL ’25},
  pages = {8912--8924},
  year = {2025},
  publisher = {Association for Computational Linguistics},
  address = {Albuquerque, New Mexico, USA},
  doi = {10.18653/v1/2025.naacl-long.449},
  url = {https://aclanthology.org/2025.naacl-long.449/}
}

@article{zhang2025whentouse,
  title   = {When to Use Graphs in Retrieval-Augmented Generation: A Comprehensive Analysis for Graph Retrieval-Augmented Generation},
  author = {Zhang, Qinggang and et al.},
  journal = {arXiv preprint arXiv:2506.05690},
  year    = {2025},
  url     = {https://arxiv.org/abs/2506.05690}
}

@InProceedings{10.1007/978-3-032-00633-2_3,
author = {Hamzic, Dzenan and et al.},
title="Enhancing Cyber Situational Awareness with AI: A Novel Pipeline Approach for Threat Intelligence Analysis and Enrichment",
booktitle="Availability, Reliability and Security",
year="2025",
publisher="Springer Nature Switzerland",
address="Cham",
pages="44--62",
isbn="978-3-032-00633-2"
}

@book{vanrijsbergen1979information,
  title={Information Retrieval},
  author={van Rijsbergen, C. J.},
  year={1979},
  publisher={Butterworth-Heinemann}
}

@inproceedings{yang2018hotpotqa,
  title     = {HotpotQA: A Dataset for Diverse, Explainable Multi-hop Question Answering},
  author    = {Yang, Zhilin and Qi, Peng and Zhang, Saizheng and Bengio, Yoshua and Cohen, William W. and Salakhutdinov, Ruslan and Manning, Christopher D.},
  booktitle = {Proceedings of the 2018 Conference on Empirical Methods in Natural Language Processing},
  year      = {2018},
  publisher = {Association for Computational Linguistics},
  doi       = {10.18653/v1/D18-1259},
  url       = {https://aclanthology.org/D18-1259/}
}

@inproceedings{rajpurkar2018know,
  title     = {Know What You Don't Know: Unanswerable Questions for {SQuAD}},
  author    = {Rajpurkar, Pranav and Jia, Robin and Liang, Percy},
  booktitle = {Proceedings of the 56th Annual Meeting of the Association for Computational Linguistics (Volume 1: Long Papers)},
  year      = {2018},
  publisher = {Association for Computational Linguistics},
  doi       = {10.18653/v1/P18-2124},
  url       = {https://aclanthology.org/P18-2124/}
}

@article{johnson2017faiss,
  title        = {Billion-Scale Similarity Search with {GPUs}},
  author={Johnson, Jeff and et al.},
  journal      = {arXiv preprint arXiv:1702.08734},
  year         = {2017},
  doi          = {10.48550/arXiv.1702.08734},
  url          = {https://arxiv.org/abs/1702.08734}
}

@inproceedings{francis2018cypher,
  title     = {Cypher: An Evolving Query Language for Property Graphs},
  author = {Francis, Nadime and et al.},
  booktitle = {Proceedings of the 2018 International Conference on Management of Data (SIGMOD '18)},
  year      = {2018},
  publisher = {Association for Computing Machinery},
  doi       = {10.1145/3183713.3190657},
  url       = {https://doi.org/10.1145/3183713.3190657}
}

@misc{yao2023react,
  title        = {ReAct: Synergizing Reasoning and Acting in Language Models},
  author = {Yao, Shunyu and et al.},
  year         = {2023},
  eprint       = {2210.03629},
  archivePrefix= {arXiv},
  primaryClass = {cs.CL},
  doi          = {10.48550/arXiv.2210.03629},
  url          = {https://arxiv.org/abs/2210.03629}
}

@inproceedings{shinn2023reflexion,
  title     = {Reflexion: Language Agents with Verbal Reinforcement Learning},
  author = {Shinn, Noah and et al.},
  booktitle = {Advances in Neural Information Processing Systems},
  volume    = {36},
  year      = {2023},
  url       = {https://arxiv.org/abs/2303.11366}
}

\end{document}